\documentclass[11pt]{article}
\PassOptionsToPackage{hyperfootnotes=false}{hyperref}
\usepackage[preprint]{acl}
\usepackage{times}
\usepackage{latexsym}
\usepackage[T1]{fontenc}
\usepackage[utf8]{inputenc}
\usepackage{microtype}
\usepackage{inconsolata}
\usepackage{graphicx}
\usepackage{amsmath,amssymb,booktabs,array,multirow,tabularx,colortbl}
\usepackage{longtable}
\usepackage{xcolor,adjustbox}
\usepackage{tikz}
\definecolor{drop1}{HTML}{FFF1F0}
\definecolor{drop2}{HTML}{FDD5D0}
\definecolor{drop3}{HTML}{F5A398}
\definecolor{drop4}{HTML}{E87368}
\newsavebox{\dropscorebox}
\DeclareRobustCommand{\dropscore}[2]{%
  \sbox{\dropscorebox}{#2}%
  \makebox[\wd\dropscorebox][c]{\smash{%
    \tikz[baseline=(score.base)]{\node[fill=drop#1,rounded corners=1.2pt,
      inner xsep=1.2pt,inner ysep=0.6pt,outer sep=0pt] (score) {\usebox{\dropscorebox}};}%
  }}%
}
\newcommand{\lscore}[3]{\begin{tabular}{@{}c@{}}#1\\[-2pt]{\scriptsize\textcolor{#2}{#3}}\end{tabular}}

\title{VidToC: Rethinking Video Token Compression\\
for Multimodal Large Language Models}

\author{
Zihan Xiao\textsuperscript{1}\thanks{Equal contribution.}\quad
Boshen Xu\textsuperscript{1}\footnotemark[1]\quad
Pei Fu\textsuperscript{2}\quad
Jianzhong Ju\textsuperscript{2}\\
\bfseries Zhenbo Luo\textsuperscript{2}\quad
Jian Luan\textsuperscript{2}\quad
Qin Jin\textsuperscript{1}\thanks{Corresponding author.}\\
\textsuperscript{1}AIM3 Lab, Renmin University of China\\
\textsuperscript{2}MiLM Plus, Xiaomi Inc.
}
\begin{document}

\maketitle

\begin{abstract}
Extending multimodal large language models (MLLMs) from images to videos faces substantial memory and latency challenges due to the massive number of visual tokens.
These challenges have driven the development of numerous visual token compression methods and more efficient LLM architectures.
However, existing compression methods are often evaluated with different training setups, tasks, and models, making fair comparison difficult.
Meanwhile, the growing use of hybrid Mamba-Transformer architectures raises questions about whether compression strategies transfer across backbones.
We present \textbf{VidToC}, a controlled evaluation framework for representative compression algorithms, LLM backbones, and time encoding strategies, covering training-free and fine-tuned settings on three complementary video understanding tasks.
Our results reveal task-dependent compression behavior and characterize the effects of compression placement, time encoding, and backbone choice.
These findings provide empirical guidance for designing and selecting video token compression strategies across tasks and training settings.
\end{abstract}

\begin{figure*}[tb]
  \centering
  \includegraphics[width=0.9\linewidth]{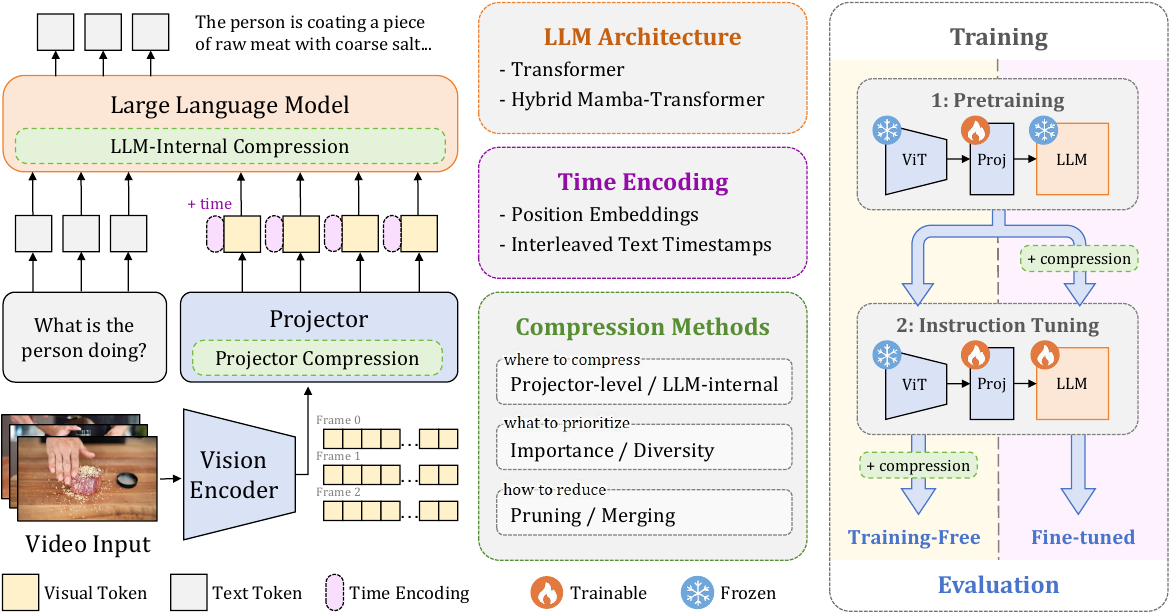}
  \caption{
  Overview of our evaluation framework VidToC.
  Left: the illustration of a typical video MLLM with possible compression modules.
  Middle: the three critical components of video compression systems, including LLM architectures, time encodings, and compression methods.
  Right: compression modules are added either in a training-free manner after instruction tuning, or in a fine-tuned manner where the model is trained with the compression module.}
  \label{fig:framework}
\end{figure*}

\section{Introduction}
\label{sec:intro}

Developing general-purpose multimodal large language models (MLLMs)~\cite{comanici2025gemini2.5pro,achiam2023gpt4v} for video understanding has attracted increasing attention, enabling applications ranging from video platform services~\cite{yang2023vidchapters} to embodied agents~\cite{black2024pi_0}.
Scaling these models to long videos exposes a fundamental bottleneck: even a one-minute short video sampled at 1 frame per second can produce over ten thousand visual tokens when encoded by a standard vision encoder~\cite{zhai2023siglip}.
Such explosive token growth significantly increases the memory footprint for KV-cache storage and inference latency caused by the quadratic attention complexity in transformer-based MLLMs.
Recent studies address this issue through two complementary directions: 1)~\emph{visual token compression methods} that reduce the number of tokens passed to the language model~\cite{fastv,vispruner,alvar2025divprune,wen2025dart}, and 2)~\emph{efficient backbone architectures} that reduce the cost of processing long sequences~\cite{gu2024mamba,dao2024mamba2,basant2025nanov2}.

However, compression methods are often evaluated with different training setups, tasks, and models, making it difficult to isolate the effect of the compression method itself.
This issue is especially important for video MLLMs, since many token-compression studies still focus primarily on image understanding and leave video-specific behavior less explored.
At the same time, recent video MLLMs~\cite{xu2025timeviper,deshmukh2025nanov2vl,xu2025auroralong} increasingly introduce linearized state-space layers~\cite{gu2024mamba,dao2024mamba2} to improve efficiency.
Validating the effectiveness of video token compression methods in hybrid Mamba-Transformer architectures and comparing their performance across different architectures has become increasingly urgent.

To address these challenges, we establish \textbf{VidToC}, a framework that systematically studies representative compression algorithms, Transformer and hybrid Mamba-Transformer backbones, and time encoding strategies.
Our evaluation covers training-free and fine-tuned settings on three complementary tasks, namely general QA, temporal grounding, and detailed captioning.

Our findings can be distilled into three insights.
(1)~In training-free settings, simple methods are competitive for general QA and detailed captioning, while temporal grounding benefits from suitable importance-based selection and preserving original position indices.
(2)~Under matched token budgets, projector-level compression is a strong training-free default, while multi-stage combinations can balance performance across tasks.
(3)~Relative method strengths broadly transfer across the two backbones, but the evaluated hybrid model exhibits greater output stability under aggressive merging and greater sensitivity to training-free compression on temporal grounding.

Our contributions are threefold:
\begin{itemize}
  \item We introduce VidToC, an extensible framework that unifies the implementation of representative token compression methods and supports their controlled evaluation across general QA, temporal grounding and detailed captioning tasks.
  \item We formulate video token compression as a joint design problem involving compression methods, LLM architectures, and time encoding strategies, and systematically characterize their interactions under training-free and fine-tuned settings.
  \item We derive practical design guidance for selecting and combining compression methods, preserving temporal information, and determining when model adaptation and backbone-specific validation become important.
\end{itemize}

\section{Related Work}
\label{sec:related}

\subsection{MLLMs for Video Understanding}
Multimodal large language models have rapidly expanded from image~\cite{liu2023llava,alayrac2022flamingo,dai2023instructblip} to video comprehension~\cite{zhang2025llavavideo,lillavaov,bai2025qwen2.5vl,bai2025qwen3vl}.
However, the linear growth of visual tokens with frame count creates a fundamental efficiency bottleneck, exacerbating the quadratic complexity of standard Transformers~\cite{vaswani2017attention}.
To address this, research now emphasizes efficient architectures and visual token compression.
Architecturally, state-space models like Mamba~\cite{gu2024mamba,dao2024mamba2} achieve linear time complexity via implicit state compression~\cite{zhao2025cobra,liao2025mmmamba,zuo2024falcon,xu2025auroralong}.
To bridge the performance gap between pure linear models and transformer-based models, hybrid architectures interleaving Mamba and attention layers~\cite{basant2025nanov2,li2025matvlm,ren2025vamba,xu2025timeviper} have gained attention.
Although these models successfully lower per-token costs, the massive volume of raw tokens remains a bottleneck.
In this work, we compare token compression across a pure Transformer and a hybrid Mamba-Transformer backbone to understand how architectural design interacts with compression.

\subsection{Visual Token Compression}
Visual token compression reduces the number of visual tokens to mitigate computational overhead.
Early parametric approaches train lightweight query-based modules to extract compact representations~\cite{alayrac2022flamingo,dai2023instructblip}, but require dedicated training and limit portability to new architectures.
To eliminate the training process, parameter-free token selection has emerged, which can be roughly divided into importance-based and diversity-based methods.
Importance-based methods~\cite{xing2024pyramiddrop,song2025trim,fastv} score tokens by their saliency.
For example, FastV~\cite{fastv} retains the visual tokens that receive the highest attention score from the last text token.
Diversity-based methods maximize visual content coverage by retaining a set of distinct tokens~\cite{alvar2025divprune,wen2025dart,bolya2022tokenmerge}.
To mitigate bias from single-criterion token selection, some methods combine importance and diversity to enhance performance~\cite{vispruner,li2025mob,yang2025visionzip}.
While parametric methods have been extensively explored, token selection strategies lack comprehensive and fair evaluation for video MLLMs.
In this work, we systematically evaluate these methods within controlled experimental setups.

\section{Method}
\label{sec:method}

We construct a controlled evaluation framework to study visual token compression for video understanding.
As illustrated in Figure~\ref{fig:framework}, we adopt a LLaVA-style~\cite{liu2023llava} architecture comprising a ViT vision encoder, a two-layer MLP projector, and an LLM.
Given a video with $T$ frames and $M$ patches per frame, the encoder produces a sequence of $N$ visual tokens, where $N = T \times M$. Then, the projector maps these tokens into the text embedding space of the LLM.
The projected visual tokens are concatenated with text tokens, and then processed by the LLM to produce output answers. Visual token compression usually occurs at the projector or LLM layers.

VidToC examines three key components of video compression systems: LLM architectures, compression methods, and time encoding strategies.
We introduce these components in order, then describe the two evaluation settings.

\begin{table*}[t]
  \centering
  \small
  \begin{minipage}{0.89\textwidth}
  \centering
  \setlength{\tabcolsep}{3pt}
  \begin{tabularx}{\linewidth}{@{}>{\raggedright\arraybackslash}p{0.34\linewidth} >{\centering\arraybackslash}p{0.06\linewidth} >{\raggedright\arraybackslash}X@{}}
    \toprule
    Method & Type & Core idea \\
    \midrule
    \rowcolor{gray!10}
    \multicolumn{3}{@{}l}{\emph{Simple methods}} \\
    Uniform Prune & P & Keep tokens at fixed intervals. \\
    Window Prune & P & Keep the center token in each spatial grid. \\
    Spatial Pool & M & Average-pool spatial tokens within each frame. \\
    Spatiotemporal Pool & M & Average-pool tokens jointly over space and time. \\
    Pixel Unshuffle~\cite{chen2024internvl2.5}
      & M & Fold spatial dimensions into channels. \\
    \midrule
    \rowcolor{gray!10}
    \multicolumn{3}{@{}l}{\emph{Complex methods}} \\
    Temporal Prune~\cite{yao2025timechat-online}
      & D,P & Remove tokens similar to the same-position token in the previous frame. \\
    FPS~\cite{alvar2025divprune,li2025mob}
      & D,P & Iteratively keep tokens farthest from selected tokens. \\
    GDR~\cite{wen2025dart}
      & D,P & Sample pivots and remove their nearest neighbors. \\
    CLS-Attn~\cite{vispruner}
      & I,P & Keep tokens with high \texttt{[CLS]} attention. \\
    CLS-Sim~\cite{chen2025recoverable}
      & I,P & Keep tokens most similar to the \texttt{[CLS]} token. \\
    Text-Attn~\cite{fastv}
      & I,P & Keep visual tokens with high final-text-token attention. \\
    Text-Sim~\cite{li2025mob}
      & I,P & Keep visual tokens most similar to text tokens. \\
    Bipartite Prune~\cite{vispruner}
      & D,P & Remove the most similar token pairs. \\
    Bipartite Merge~\cite{bolya2022tokenmerge}
      & D,M & Merge the most similar token pairs. \\
    Selective SlowFast~\cite{shen2025longvu}
      & I,M & Allocate higher resolution to text-relevant frames. \\
    \bottomrule
  \end{tabularx}
  \caption{Atomic compression methods evaluated in this study. 
  Tags: I = importance-based, D = diversity-based, P = pruning, M = merging.}
  \label{tab:compression_methods}
  \end{minipage}
\end{table*}

\subsection{LLM Architectures}
\label{subsec:archs}

The internal sequence modeling mechanism of the LLM backbone can influence how well compressed tokens are utilized.
To study this interaction, we evaluate two representative LLM architectures: a pure Transformer backbone and a hybrid Mamba-Transformer design.
\textbf{Transformers} remain the dominant architecture for multimodal reasoning due to their ability to capture complex global dependencies through dense token interactions.
However, the quadratic complexity of self-attention also makes them particularly sensitive to long token sequences and aggressive token reduction.
As a representative Transformer backbone, we adopt Qwen2.5-7B-Base~\cite{qwen2.5} (Qwen), a 28-layer Transformer model employing grouped-query attention where each layer performs full self-attention over the entire token sequence.
\textbf{Hybrid Mamba-Transformers} aim to combine the high computational efficiency of linear sequence models with the strong reasoning capabilities of attention mechanisms.
They are particularly adept at processing extremely long sequences with minimal memory overhead while maintaining competitive reasoning performance.
To represent this paradigm, we adopt Nemotron-Nano-9B-V2-Base~\cite{basant2025nanov2} (Nano), which integrates 27 Mamba-2 layers, 25 MLP layers, and 4 evenly distributed attention layers.
By evaluating token compression strategies across these two backbone architectures, our framework allows us to analyze whether compression behavior depends on the underlying sequence modeling paradigm.

\subsection{Compression Methods}
\label{subsec:compression_methods}
We decompose visual token compression methods into three dimensions:
where compression is applied in the pipeline, what tokens are prioritized, and how the reduction operation is performed.
Table~\ref{tab:compression_methods} lists the atomic compression methods considered in our study, which are extracted from existing representative works.
We also evaluate some combined variants constructed by mixing or sequentially applying these methods.

\noindent\textbf{Where to compress.}
Visual token compression can occur at projector or LLM layers.
\emph{Projector-level} compression~\cite{vispruner,alvar2025divprune,yao2025timechat-online} reduces visual tokens before the LLM. LLM-internal compression~\cite{fastv,xing2024pyramiddrop,wen2025dart,li2025mob} acts between LLM decoder layers, allowing the model to leverage intermediate attention signals and perform cross-modal interaction before discarding tokens.
However, \emph{LLM-internal} compression incurs the cost of uncompressed early layers.
At the same retention rate, LLM-internal compression processes more visual tokens across layers than projector-level compression.
To relieve this bias, we match the average number of visual tokens per layer.
Let $N$ be the number of visual tokens, $L$ the number of LLM layers, $r$ the projector-level retention rate, and $r'$ the LLM-internal retention rate after layer $l$. 
We choose $r'$ by matching $NrL = Nl + Nr'(L-l)$.
More discussion on this matching strategy and efficiency comparisons is provided in Appendices~\ref{app:matched_budget} and~\ref{app:efficiency}.

\noindent\textbf{What to prioritize.}
Token selection criteria are commonly categorized into importance-based and diversity-based methods.
\emph{Importance-based} methods prioritize the most salient visual tokens, typically by \texttt{[CLS]} token attention~\cite{yang2025visionzip,vispruner}, text-to-vision attention at decoder layer~\cite{fastv,xing2024pyramiddrop}, or visual-text similarity~\cite{shen2025longvu,li2025mob}, but may overlook less salient but informative tokens.
In contrast, \emph{diversity-based} methods select tokens that are mutually dissimilar~\cite{alvar2025divprune,li2025mob,bolya2022tokenmerge,vispruner}, preserving a wider range of visual information.
Some methods further combine the two principles by allocating separate selection budgets or computing hybrid importance-diversity scores~\cite{li2025mob,vispruner,yang2025visionzip}.

\noindent\textbf{How to reduce.}
After token selection, the actual token reduction can be performed by directly pruning the unselected tokens or merging them into new representations.
\emph{Pruning}~\cite{alvar2025divprune,vispruner,wen2025dart} cannot increase per-token information density, which may lead to irreversible loss of potentially important visual details under high compression rates.
\emph{Merging}~\cite{bolya2022tokenmerge,yang2025visionzip,zhang2024sparsevlm} increases per-token information density, but introduces newly synthesized token representations that may lie outside the original training distribution of the LLM, leading to performance degradation in training-free settings.
\subsection{Time Encoding}
\label{subsec:time_encoding}
Unlike static images, videos possess an essential temporal dimension.
How to preserve temporal information during compression is thus a critical design choice.
Temporal information is typically encoded implicitly through position embeddings or explicitly through text timestamps.

\noindent\textbf{Position embeddings.}
By default, Qwen uses 1D-RoPE~\cite{su2024rope}, whereas Nano has no explicit positional embeddings (NoPE) because its recurrent state-space architecture inherently models sequence order.
We also evaluate MRoPE-interleave (MRoPE-I)~\cite{bai2025qwen3vl}, which decomposes position embeddings into temporal, height, and width components to explicitly capture the 3D structure of videos.
Another key design question is how to assign position indices to the compressed tokens.
The \emph{preserve} strategy keeps the original indices of the surviving tokens, leaving numerical gaps where tokens were discarded or merged.
The \emph{reset} strategy instead reassigns indices to form a new continuous sequence after compression.

\noindent\textbf{Interleaved text timestamps.}
Inserting text timestamps before each frame provides explicit temporal cues that are independent of the position embedding scheme and unaffected by compression. Recent studies~\cite{bai2025qwen3vl,zhang2025timelens} have shown that it can effectively improve long video understanding.

\subsection{Evaluation Settings}
\label{subsec:training_procedure}
We adopt a standard two-stage training pipeline for vision-language models.
During the pretraining stage, we train the projector while keeping both the vision encoder and the LLM frozen, allowing the model to align visual representations with the language embedding space.
During instruction tuning stage, we jointly fine-tune the projector and the LLM on diverse image and video instruction data, enabling the model to perform instruction-following and a variety of downstream multimodal reasoning tasks.

To systematically evaluate the impact of visual token compression, each method is tested under two different settings.
In the \textbf{training-free} setting, the compression module is inserted into a fully trained uncompressed model at inference time, and performance is evaluated directly without any additional training.
This setting measures the intrinsic compatibility of a compression strategy with existing pretrained models.
In the \textbf{fine-tuned} setting, the compression module is introduced during the instruction-tuning stage, and the model is fine-tuned end-to-end before evaluation.
This allows the model to adapt to the distribution shift introduced by token compression and potentially recover performance losses.
Comparing these two settings provides insight into whether compression methods are inherently robust or require adaptation through additional training.

\begin{table*}[t]
  \centering
  \small
  \begin{adjustbox}{max width=\textwidth}
  \begin{tabular}{@{}l|cc|cc|cc|cc|cc|cc@{}}
    \toprule
    \multirow{3}{*}{Method} &
    \multicolumn{6}{c|}{Qwen2.5-7B} &
    \multicolumn{6}{c}{Nano-9B-v2} \\
    \cmidrule(l{1pt}r{1pt}){2-7}
    \cmidrule(l{1pt}r{1pt}){8-13}
    & \multicolumn{2}{c|}{VideoMME}
    & \multicolumn{2}{c|}{Charades}
    & \multicolumn{2}{c|}{VDC}
    & \multicolumn{2}{c|}{VideoMME}
    & \multicolumn{2}{c|}{Charades}
    & \multicolumn{2}{c}{VDC} \\
    & TF & FT & TF & FT & TF & FT
    & TF & FT & TF & FT & TF & FT \\
    \midrule
    Uncompressed
      & \multicolumn{2}{c|}{60.19} & \multicolumn{2}{c|}{33.83} & \multicolumn{2}{c|}{44.74}
      & \multicolumn{2}{c|}{57.89} & \multicolumn{2}{c|}{32.50} & \multicolumn{2}{c}{44.48} \\
    \midrule
    \rowcolor{gray!10}
    \multicolumn{13}{@{}l}{\emph{Simple methods}} \\
    Uniform Prune
      & \dropscore{2}{54.07} & \dropscore{1}{56.22} & \dropscore{4}{17.29} & \dropscore{2}{27.85} & \dropscore{1}{\textbf{42.20}} & 43.19
      & \dropscore{1}{\textbf{53.63}} & \dropscore{1}{54.41} & \dropscore{4}{14.62} & \dropscore{2}{27.10} & \textbf{42.48} & 43.42 \\
    Window Prune
      & \dropscore{1}{\underline{54.70}} & 57.56 & \dropscore{4}{18.08} & \dropscore{2}{29.68} & \dropscore{1}{40.58} & 43.38
      & \dropscore{1}{53.30} & \dropscore{1}{54.63} & \dropscore{4}{14.10} & \dropscore{2}{28.72} & \dropscore{2}{39.64} & 42.46 \\
    Spatial Pool
      & \dropscore{4}{17.96} & \underline{58.19} & \dropscore{4}{16.04} & \dropscore{1}{31.05} & \dropscore{4}{14.12} & 43.52
      & \dropscore{2}{51.93} & 55.59 & \dropscore{4}{13.93} & \dropscore{1}{29.74} & \dropscore{2}{37.13} & 42.95 \\
    Spatiotemporal Pool
      & \dropscore{2}{51.56} & 57.48 & \dropscore{4}{16.90} & \dropscore{1}{30.90} & \dropscore{2}{38.18} & 44.13
      & \dropscore{2}{51.67} & \underline{56.04} & \dropscore{4}{14.16} & \dropscore{1}{29.76} & \dropscore{2}{39.19} & 43.24 \\
    Pixel Unshuffle
      & -- & \dropscore{1}{56.33} & -- & \dropscore{1}{\underline{31.27}} & -- & 43.93
      & -- & \textbf{56.63} & -- & \dropscore{1}{\textbf{30.86}} & -- & 43.41 \\
    \midrule
    \rowcolor{gray!10}
    \multicolumn{13}{@{}l}{\emph{Atomic methods}} \\
    CLS-Attn
      & \dropscore{2}{51.74} & \dropscore{1}{55.37} & \dropscore{2}{\underline{29.21}} & \dropscore{2}{30.24} & \dropscore{2}{36.64} & \dropscore{1}{41.20}
      & \dropscore{2}{51.89} & 55.78 & \dropscore{4}{\underline{15.88}} & \dropscore{1}{29.67} & \dropscore{2}{38.30} & \dropscore{1}{41.70} \\
    CLS-Sim
      & \dropscore{2}{51.19} & \dropscore{1}{55.59} & \dropscore{4}{18.12} & \dropscore{2}{27.15} & \dropscore{3}{34.47} & \dropscore{1}{41.41}
      & \dropscore{2}{49.15} & \dropscore{1}{53.19} & \dropscore{4}{13.93} & \dropscore{2}{26.19} & \dropscore{3}{33.79} & \dropscore{2}{39.84} \\
    Bipartite Prune
      & \dropscore{2}{53.37} & \dropscore{1}{55.19} & \dropscore{3}{25.16} & \dropscore{1}{30.60} & \dropscore{1}{\underline{42.00}} & 44.29
      & \dropscore{1}{52.33} & \dropscore{1}{53.85} & \dropscore{4}{14.16} & \dropscore{2}{28.17} & \dropscore{1}{42.13} & \underline{43.46} \\
    Bipartite Merge
      & \dropscore{2}{53.37} & 58.11 & \dropscore{3}{26.64} & \dropscore{2}{30.08} & \dropscore{2}{40.04} & \textbf{44.74}
      & \dropscore{1}{53.30} & \dropscore{1}{54.44} & \dropscore{4}{14.75} & \dropscore{1}{30.28} & \dropscore{1}{41.58} & \textbf{44.45} \\
    Temporal Prune$^\dagger$
      & \dropscore{1}{\textbf{54.96}} & 57.37 & \dropscore{4}{22.31} & \dropscore{2}{28.98} & \dropscore{1}{41.96} & 43.30
      & \dropscore{1}{53.37} & \dropscore{1}{53.56} & \dropscore{4}{14.86} & \dropscore{3}{25.75} & \dropscore{1}{42.10} & 43.10 \\
    Selective SlowFast$^\dagger$
      & \dropscore{2}{52.70} & \dropscore{1}{55.59} & \dropscore{4}{17.08} & \dropscore{3}{27.01} & \dropscore{1}{41.04} & 43.05
      & \dropscore{2}{51.67} & \dropscore{1}{53.81} & \dropscore{4}{14.05} & \dropscore{3}{25.97} & \dropscore{1}{40.82} & 43.04 \\
    GDR$^\dagger$
      & \dropscore{2}{53.78} & \dropscore{1}{55.30} & \dropscore{2}{28.66} & \dropscore{1}{30.53} & \dropscore{2}{40.13} & 42.94
      & \dropscore{1}{\underline{53.41}} & \dropscore{1}{54.44} & \dropscore{4}{\textbf{16.79}} & \dropscore{1}{\underline{30.48}} & \underline{42.36} & \dropscore{1}{41.61} \\
    FPS$^\dagger$
      & \dropscore{1}{54.41} & \textbf{58.56} & \dropscore{4}{20.04} & \dropscore{1}{\textbf{31.95}} & \dropscore{1}{41.57} & 44.21
      & \dropscore{1}{52.85} & \dropscore{1}{54.70} & \dropscore{4}{14.70} & \dropscore{2}{28.72} & 42.28 & 43.18 \\
    Text-Attn$^\dagger$
      & \dropscore{2}{51.70} & \dropscore{1}{55.44} & \dropscore{2}{28.68} & \dropscore{2}{29.63} & \dropscore{3}{35.16} & \dropscore{1}{41.69}
      & -- & -- & -- & -- & -- & -- \\
    Text-Sim$^\dagger$
      & \dropscore{2}{49.37} & \dropscore{1}{55.37} & \dropscore{4}{16.89} & \dropscore{2}{27.53} & \dropscore{4}{30.90} & \dropscore{1}{41.47}
      & \dropscore{2}{49.37} & \dropscore{2}{51.63} & \dropscore{4}{13.98} & \dropscore{2}{26.24} & \dropscore{3}{34.32} & \dropscore{1}{41.10} \\
    \midrule
    \rowcolor{gray!10}
    \multicolumn{13}{@{}l}{\emph{Mixed methods}} \\
    CLS-Attn+Bip. Prune
      & \dropscore{2}{52.78} & \dropscore{1}{56.15} & \dropscore{3}{24.56} & \dropscore{2}{29.04} & \dropscore{1}{40.27} & 43.53
      & \dropscore{2}{51.63} & 55.07 & \dropscore{4}{15.68} & \dropscore{2}{29.06} & \dropscore{1}{40.43} & 43.14 \\
    Text-Sim+FPS$^\dagger$
      & \dropscore{2}{53.81} & 57.67 & \dropscore{4}{20.25} & \dropscore{1}{31.01} & \dropscore{2}{39.58} & 44.19
      & \dropscore{1}{52.22} & \dropscore{1}{54.11} & \dropscore{4}{14.45} & \dropscore{2}{28.96} & \dropscore{1}{41.11} & 42.60 \\
    Text-Attn+FPS$^\dagger$
      & \dropscore{2}{52.93} & \dropscore{1}{56.41} & \dropscore{2}{\textbf{29.30}} & \dropscore{1}{30.52} & \dropscore{1}{40.50} & \underline{44.32}
      & -- & -- & -- & -- & -- & -- \\
    \bottomrule
  \end{tabular}
  \end{adjustbox}
  \caption{
    Compression results for Qwen2.5-7B and Nano-9B-v2. TF: training-free; FT: fine-tuned. $\dagger$ indicates methods that may produce a variable number of tokens per frame. Text-Attn results are unavailable for Nano-9B-v2 because its attention layers are too deep to satisfy the matched token budget. For mixed methods, each component uses half of the token budget.
    Color indicate relative score drops from Uncompressed in five ranges $[0,5)\%$, \dropscore{1}{$[5,10)\%$}, \dropscore{2}{$[10,20)\%$}, \dropscore{3}{$[20,30)\%$}, and \dropscore{4}{$[30,+\infty)\%$}.
  }
  \label{tab:results_main}
\end{table*}

\section{Experiments}
\label{sec:experiments}

This section first describes our experimental setup, including implementation details and evaluation benchmarks.
We then discuss results on task-dependent compression behavior, compression position, time encoding and backbone architecture.

\subsection{Experiment Setup}
\label{subsec:setup}

\noindent\textbf{Implementation details.}
We employ SigLIP-SO400M~\cite{zhai2023siglip} as vision encoder and a two-layer GELU MLP as the projector.
During the pretraining stage, we utilize 3 million CC12M images~\cite{changpinyo2021cc12m} paired with PixelProse captions~\cite{singla2024pixelprose}.
During visual instruction tuning, we train on 1.3M video-centric instructions primarily derived from LLaVA-Video~\cite{zhang2025llavavideo}, 2.8M image-centric instructions from LLaVA-OneVision~\cite{lillavaov}, together with 26K detailed captioning samples~\cite{xu2025timeviper} and 250K temporal grounding examples~\cite{xu2025timeviper}.
Videos are sampled at 1 frame per second with a maximum frame budget. When the sampled frames exceed this budget, we uniformly subsample frames from the entire video. The default maximum frame budget is 32 frames.
Interleaved text timestamps are formatted as \texttt{<0.0 seconds>} and prepended to each frame.
Projector-level compression is fixed at a retention rate of $1/16$. For LLM-internal compression methods, we adjust the final retention rate to ensure a matched token budget, as described in Section~\ref{subsec:compression_methods}.

\noindent\textbf{Benchmarks.}
We evaluate methods on three complementary benchmarks of video understanding.
(1)~\textbf{VideoMME}~\cite{fu2025videomme} evaluates general QA through multiple-choice video questions.
  Given a video and a question with several candidate answers, the model must select the correct option.
  The benchmark includes 2.7K QA samples covering short, medium, and long videos across 12 subcategories.
(2)~\textbf{Charades}~\cite{gao2017charades,zhang2025timelens} focuses on temporal grounding.
  Given a natural language query, the model must localize the corresponding temporal segment within the video. We use the TimeLens relabeled version. The benchmark contains 3.3K natural language queries over 1.3K indoor activity videos.
(3)~\textbf{VDC}~\cite{chai2025auroracap} evaluates detailed captioning.
  Given a video sequence, the model must generate fine-grained, comprehensive descriptive paragraphs. The benchmark consists of 1.0K videos, and is evaluated via query-conditioned scoring using LLaMA3.1-8B~\cite{dubey2024llama3}.

We report overall accuracy for VideoMME and VDC, and mean IoU for Charades. VideoMME is evaluated without subtitles.

\subsection{Task-Dependent Compression Behavior}

Table~\ref{tab:results_main} shows that both task type and training setting can change the relative behavior of compression methods. In training-free settings, the behavior patterns of compression methods across tasks are more diverse and observable.

\noindent\textbf{Simple methods are competitive for general QA and detailed captioning, but not for temporal grounding.}
In training-free evaluations, simple methods are competitive for general QA and detailed captioning.
Uniform Prune achieves the best training-free VDC score on both backbones and the best VideoMME score on Nano.
Window Prune (54.70) is comparable to the strongest Qwen method, Temporal Prune (54.96), on VideoMME.
However, this trend breaks on the temporal grounding task on Charades.
In training-free evaluations on the Qwen backbone, only a small subset of methods, such as CLS-Attn and Text-Attn+FPS, achieve strong temporal grounding performance, outperforming both simple baselines and many complex approaches.
This suggests that temporal grounding specifically demands more fine-grained action cues, which only a few methods effectively capture.

\begin{figure}[t]
\centering
\includegraphics[width=\linewidth]{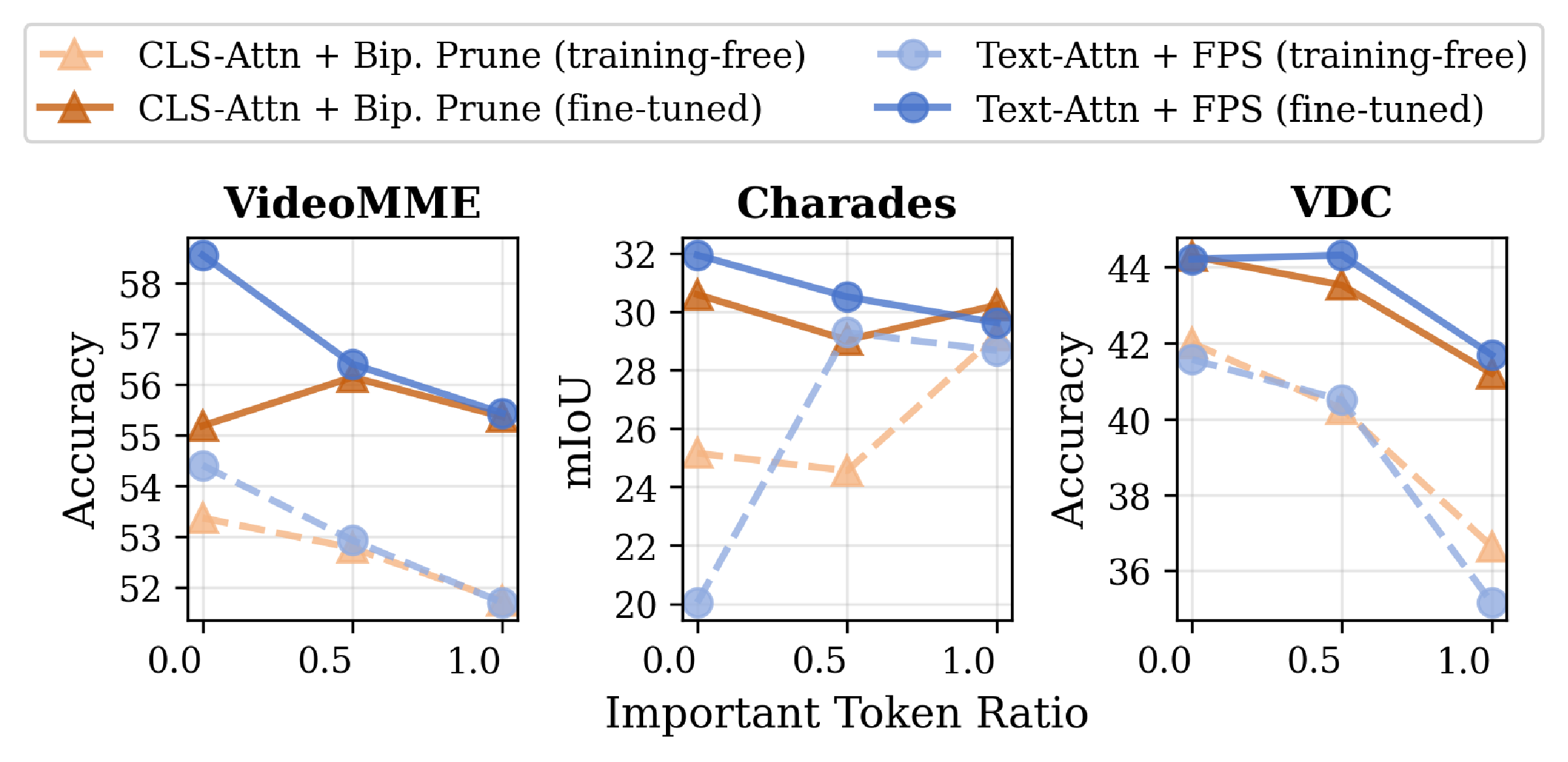}
\caption{Impact of the importance-to-diversity ratio on Qwen2.5-7B.}
\label{fig:div_imp}
\end{figure}

\noindent\textbf{Temporal grounding favors importance, while detailed captioning favors diversity.}
Figure~\ref{fig:div_imp} further compares token-selection preferences across tasks. We consider two method families combining importance-based and diversity-based selection, with importance ratios ranging from 0 (pure diversity) to 1 (pure importance).
Training-free Charades generally benefits from adding importance, although the best point can occur at a mixed ratio rather than pure importance.
In contrast, VDC decreases monotonically as the importance fraction increases for both mixtures, while VideoMME shows a milder decline.
These trends suggest different task requirements for salient content and broad coverage, making it challenging for a single configuration to serve all tasks.

\noindent\textbf{Fine-tuning generally reduces these differences.}
After instruction tuning, differences among methods narrow across all tasks on both backbones, except for temporal grounding on Nano.
Spatial Pool becomes competitive on Qwen VideoMME, reaching 58.19 compared with 58.56 for FPS.
Fine-tuning also generally reduces sensitivity to the importance fraction in Figure~\ref{fig:div_imp}.
As for temporal grounding on Nano, differences among methods widen after fine-tuning.
This may be because compressed Nano models have uniformly low grounding scores in the training-free setting, making differences among compression methods less apparent.

\subsection{Compression Position}

\begin{table}[t]
\centering
\small
\setlength{\tabcolsep}{2.3pt}
\begin{adjustbox}{max width=\columnwidth}
\begin{tabular}{@{}l|cc|cc|cc|cc@{}}
\toprule
\multirow{2}{*}{Method} & \multicolumn{2}{c|}{VideoMME} & \multicolumn{2}{c|}{Charades} & \multicolumn{2}{c|}{VDC} & \multirow{2}{*}{\shortstack{Prefill\\Time$\downarrow$}} & \multirow{2}{*}{\shortstack{Peak\\Memory$\downarrow$}} \\
& TF & FT & TF & FT & TF & FT & & \\
\midrule
\rowcolor{gray!10}
\multicolumn{9}{@{}l}{\emph{Qwen2.5-7B, L1}} \\
Uniform & \lscore{52.63}{red!75!black}{-1.44} & \lscore{56.19}{red!75!black}{-0.03} & \lscore{17.12}{red!75!black}{-0.17} & \lscore{29.71}{green!45!black}{+1.86} & \lscore{41.25}{red!75!black}{-0.95} & \lscore{43.84}{green!45!black}{+0.65} & +9.1\% & +5.3\% \\[2pt]
FPS & \lscore{52.33}{red!75!black}{-2.08} & \lscore{56.96}{red!75!black}{-1.60} & \lscore{15.95}{red!75!black}{-4.09} & \lscore{30.72}{red!75!black}{-1.23} & \lscore{39.34}{red!75!black}{-2.23} & \lscore{44.22}{green!45!black}{+0.01} & -40.4\% & +4.6\% \\[2pt]
Text-Sim & \lscore{46.07}{red!75!black}{-3.30} & \lscore{53.37}{red!75!black}{-2.00} & \lscore{15.02}{red!75!black}{-1.87} & \lscore{28.22}{green!45!black}{+0.69} & \lscore{23.04}{red!75!black}{-7.86} & \lscore{42.22}{green!45!black}{+0.75} & +11.1\% & +5.3\% \\[2pt]
Text-Attn & \lscore{50.93}{red!75!black}{-0.77} & \lscore{56.67}{green!45!black}{+1.23} & \lscore{28.48}{red!75!black}{-0.20} & \lscore{30.78}{green!45!black}{+1.15} & \lscore{35.34}{green!45!black}{+0.18} & \lscore{43.54}{green!45!black}{+1.85} & +5.4\% & +3.7\% \\[2pt]
\midrule
\rowcolor{gray!10}
\multicolumn{9}{@{}l}{\emph{Nano-9B-v2, L2}} \\
Uniform & \lscore{51.04}{red!75!black}{-2.59} & \lscore{53.37}{red!75!black}{-1.04} & \lscore{13.95}{red!75!black}{-0.67} & \lscore{27.87}{green!45!black}{+0.77} & \lscore{41.59}{red!75!black}{-0.89} & \lscore{43.44}{green!45!black}{+0.02} & +1.2\% & +3.4\% \\[2pt]
FPS & \lscore{50.33}{red!75!black}{-2.52} & \lscore{53.04}{red!75!black}{-1.66} & \lscore{14.04}{red!75!black}{-0.66} & \lscore{27.35}{red!75!black}{-1.37} & \lscore{39.80}{red!75!black}{-2.48} & \lscore{43.40}{green!45!black}{+0.22} & -39.1\% & +3.4\% \\[2pt]
Text-Sim & \lscore{46.04}{red!75!black}{-3.33} & \lscore{49.00}{red!75!black}{-2.63} & \lscore{13.93}{red!75!black}{-0.05} & \lscore{20.73}{red!75!black}{-5.51} & \lscore{32.12}{red!75!black}{-2.20} & \lscore{38.03}{red!75!black}{-3.07} & -0.1\% & +3.4\% \\[2pt]
\bottomrule
\end{tabular}
\end{adjustbox}
\caption{Comparison of LLM-internal and projector-level compression under matched token budgets. L1, L2 suggest compression occurs after the first or second decoder layer. Small colored values and efficiency metrics show the changes from Projector-level results.}
\label{tab:compression_position}
\end{table}

\noindent\textbf{Projector-level compression is a strong training-free default.}
Moving compression from the projector to L1 on Qwen or L2 on Nano reduces training-free performance in nearly every comparison in Table~\ref{tab:compression_position}.
Fine-tuning mitigates many of these losses, and some configurations outperform their projector-level counterparts.
Text-Attn shows the mildest training-free degradation and the strongest overall fine-tuned advantage over Proj, gaining 1.23, 1.15, and 1.85 points on the three tasks.
A possible explanation is that text attention can exploit early cross-modal interactions, especially after adaptation, whereas other selection criteria benefit less from these interactions while facing a lower retention rate to match the token budget.
Appendix~\ref{app:position_8x} shows that this overall trend persists at a milder $8\times$ compression rate, where compression can be inserted after deeper LLM layers.
Under our token budget matching protocol, LLM-internal compression has broadly comparable efficiency to projector-level compression, typically with slightly higher latency and peak memory.
FPS is an exception because its iterative selection requires one round per retained token, so the lower LLM-internal retention rate reduces the number of costly sequential rounds.
Appendix~\ref{app:efficiency} reports the full efficiency results and provides a detailed analysis of this behavior.

\begin{table}[t]
\centering
\small
\setlength{\tabcolsep}{2pt}
\begin{adjustbox}{max width=\columnwidth}
\begin{tabular}{@{}l|c|cc|cc|cc@{}}
\toprule
\multirow{2}{*}{Method}
& \multirow{2}{*}{Position}
& \multicolumn{2}{c|}{VideoMME}
& \multicolumn{2}{c|}{Charades}
& \multicolumn{2}{c}{VDC} \\
& & TF & FT & TF & FT & TF & FT \\
\midrule

CLS-Attn$\to$FPS & Proj,L2
& 56.81 & 57.96 & 22.05 & 32.23 & 42.30 & 43.94 \\
\textcolor{gray}{$\Delta$ vs CLS-Attn}
& \textcolor{gray}{Proj}
& \textcolor{green!45!black}{+5.07}
& \textcolor{green!45!black}{+2.59}
& \textcolor{red!75!black}{-7.16}
& \textcolor{green!45!black}{+1.99}
& \textcolor{green!45!black}{+5.66}
& \textcolor{green!45!black}{+2.74} \\
\textcolor{gray}{$\Delta$ vs FPS}
& \textcolor{gray}{Proj}
& \textcolor{green!45!black}{+2.40}
& \textcolor{red!75!black}{-0.60}
& \textcolor{green!45!black}{+2.01}
& \textcolor{green!45!black}{+0.28}
& \textcolor{green!45!black}{+0.73}
& \textcolor{red!75!black}{-0.27} \\
\midrule

Temp. Prune$\to$Text-Attn & Proj,L2
& 55.30 & 55.96 & 27.05 & 30.99 & 41.26 & 44.17 \\
\textcolor{gray}{$\Delta$ vs Temp. Prune}
& \textcolor{gray}{Proj}
& \textcolor{green!45!black}{+0.34}
& \textcolor{red!75!black}{-1.41}
& \textcolor{green!45!black}{+4.74}
& \textcolor{green!45!black}{+2.01}
& \textcolor{red!75!black}{-0.70}
& \textcolor{green!45!black}{+0.87} \\
\textcolor{gray}{$\Delta$ vs Text-Attn}
& \textcolor{gray}{Proj}
& \textcolor{green!45!black}{+3.60}
& \textcolor{green!45!black}{+0.52}
& \textcolor{red!75!black}{-1.63}
& \textcolor{green!45!black}{+1.36}
& \textcolor{green!45!black}{+6.10}
& \textcolor{green!45!black}{+2.48} \\
\midrule

Spatial Pool$\to$Text-Attn & Proj,L2
& 54.63 & 56.11 & 21.53 & 30.82 & 39.68 & 43.42 \\
\textcolor{gray}{$\Delta$ vs Spatial Pool}
& \textcolor{gray}{Proj}
& \textcolor{green!45!black}{+36.67}
& \textcolor{red!75!black}{-2.08}
& \textcolor{green!45!black}{+5.49}
& \textcolor{red!75!black}{-0.23}
& \textcolor{green!45!black}{+25.56}
& \textcolor{red!75!black}{-0.10} \\
\textcolor{gray}{$\Delta$ vs Text-Attn}
& \textcolor{gray}{Proj}
& \textcolor{green!45!black}{+2.93}
& \textcolor{green!45!black}{+0.67}
& \textcolor{red!75!black}{-7.15}
& \textcolor{green!45!black}{+1.19}
& \textcolor{green!45!black}{+4.52}
& \textcolor{green!45!black}{+1.73} \\

\bottomrule
\end{tabular}
\end{adjustbox}
\caption{
Two-stage compression on Qwen2.5-7B.
The first method operates at projector and the second after the second decoder layer.
Differences are computed against each component method applied alone at projector.
}
\label{tab:multi_stage_prune}
\end{table}

\noindent\textbf{Combining compression methods can balance task preferences.}
Table~\ref{tab:multi_stage_prune} evaluates three combinations that compress at Proj and then L2 under our token-budget matching protocol.
In training-free evaluations, Temporal Prune$\to$Text-Attn balances the stronger grounding performance of Text-Attn with the stronger captioning performance of Temporal Prune.
Combinations can also outperform both components, as CLS-Attn$\to$FPS does on training-free VideoMME and Temporal Prune$\to$Text-Attn does on Charades and VDC after fine-tuning.
However, gains depend on the training setting.
Spatial Pool$\to$Text-Attn substantially improves over Spatial Pool before fine-tuning but falls slightly below it afterward.
These results suggest that combining methods can balance task performance and sometimes exceed both single-method baselines, but still need careful configuration.

\subsection{Time Encoding}

\begin{figure}[t]
\centering
\includegraphics[width=0.9\linewidth]{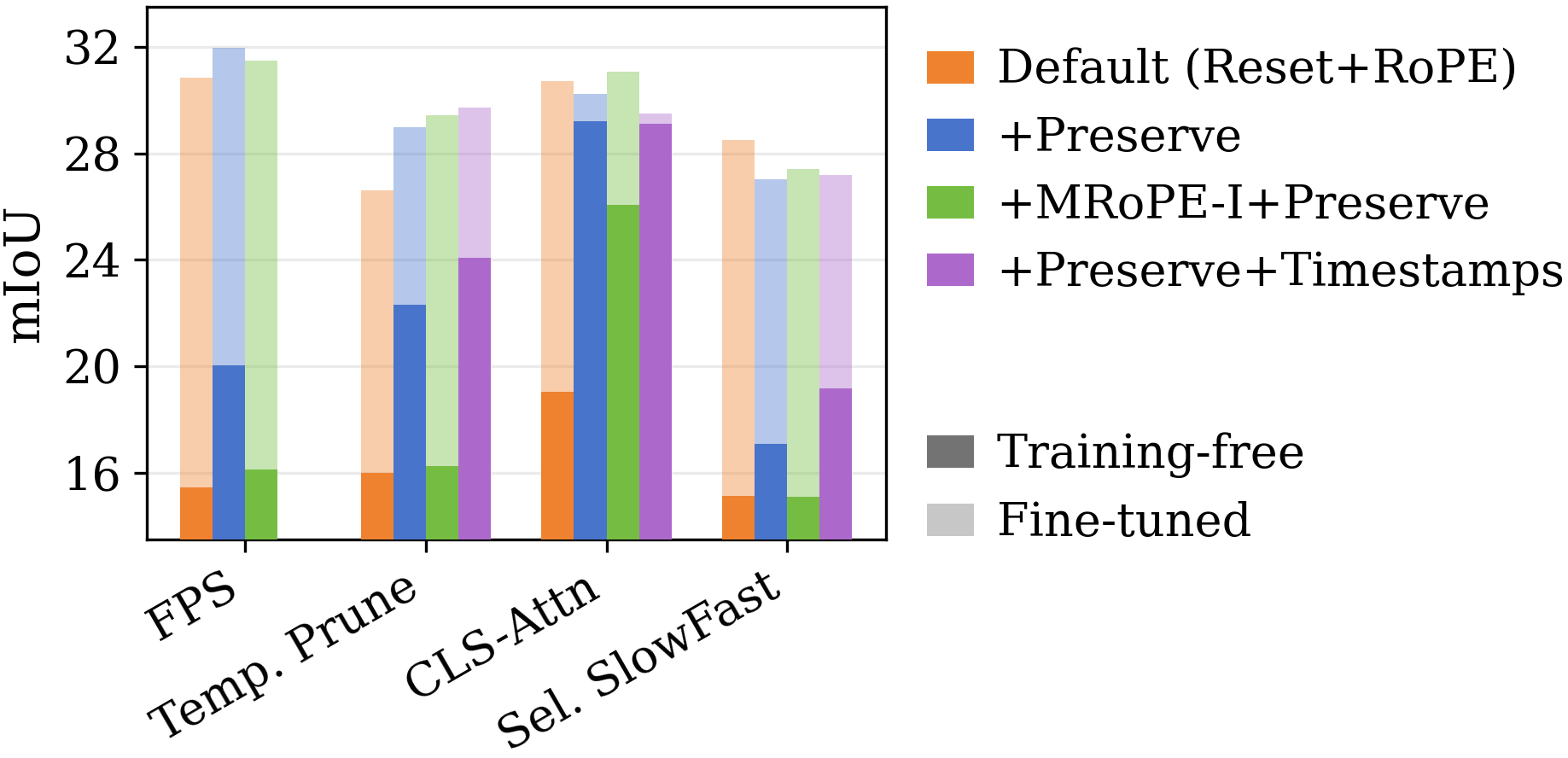}
\caption{Impact of time encoding strategies on Charades with Qwen2.5-7B.}
\label{fig:time_encoding}
\end{figure}

Figure~\ref{fig:time_encoding} compares different time encoding strategies on Charades with Qwen2.5-7B. Full results on both backbones and all three tasks are given in Appendix~\ref{app:time_encoding}.

\noindent\textbf{Preserving original position indices improves training-free temporal grounding.}
In training-free evaluations, \emph{preserve} (blue) improves Charades performance over \emph{reset} (orange) for all methods, with gains reaching up to 10 points for CLS-Attn.
After fine-tuning, the effect becomes method-dependent.
\emph{Preserve} remains advantageous for FPS and Temporal Prune, but becomes worse than \emph{reset} for CLS-Attn and Selective SlowFast.
We hypothesize that CLS-Attn produces a fixed number of tokens per frame, so under \emph{reset} the model can implicitly learn a new temporal coordinate system during fine-tuning.
Meanwhile, \emph{preserve} introduces sparse position indices, which can increase learning difficulty and reduce its advantage after fine-tuning.
For methods with a variable number of tokens per frame such as FPS and Temporal Prune, learning the implicit mapping between position embeddings and absolute time becomes challenging, so the benefits of \emph{preserve} persist even after fine-tuning.
Selective SlowFast, though with variable token counts, has only two resolution types, which may explain why \emph{reset} can be preferable after fine-tuning.

\noindent\textbf{Text timestamps mainly benefit training-free temporal grounding.}
Comparing \texttt{+Preserve} (blue) and \texttt{+Preserve+Timestamps} (purple) bars, we find that adding text timestamps further improves training-free Charades performance.
These timestamps provide explicit frame-time information that remains available after token compression.
The benefit becomes much smaller after fine-tuning, suggesting that explicit temporal cues are most useful in the training-free setting.

\noindent\textbf{MRoPE-I offers little advantage.}
MRoPE-I (green) underperforms the corresponding RoPE (blue) in most training-free evaluations, and only bring marginal advantages after fine-tuning.
This may be due to its more complex multidimensional encoding, which is hard for models to understand the sparse positions after compression.

\subsection{Backbone Architecture Comparison}

\noindent\textbf{Relative method strengths largely transfer across backbones.}
Figure~\ref{fig:method_trends_app} shows broadly similar patterns of stronger and weaker methods on Qwen and Nano in both training settings, although exact rankings vary.
For example, Uniform Prune leads training-free VDC on both backbones, while Bipartite Merge leads after fine-tuning.
CLS-Sim and Text-Sim remain relatively weak on VDC in both settings.

\noindent\textbf{Nano performs worse on training-free temporal grounding after compression.}
Although their uncompressed Charades scores are similar, Nano scores lower than Qwen for every shared compression method in Table~\ref{tab:results_main}.
For example, CLS-Attn achieves 15.88 on Nano versus 29.21 on Qwen.
The gap persists when both backbones use RoPE and preserved position indices, as shown in Appendix~\ref{app:time_encoding}.
These results motivate compression strategies that better preserve temporal information for training-free grounding on Nano.

\noindent\textbf{Nano is more robust to training-free merging.}
Under Spatial Pool, Qwen exhibits an unexpected failure mode, frequently generating corrupted text and dropping to 17.96 on VideoMME and 14.12 on VDC, while Nano shows no similar failure.
The problem recurs with another Spatial Pool seed and also appears, though much less frequently, with Bipartite Merge at $50\times$ compression (Appendix~\ref{app:output_stability}).
One possible explanation is that aggressive merging yields token representations unlike those Qwen encountered during training, whereas Nano may better tolerate them because its Mamba layers  inherently summarize sequences into fixed-size states.
Bipartite Merge only merges similar tokens, and thus introduces a smaller representational shift and largely avoids this failure.

\section{Conclusion}
\label{sec:conclusion}
We present VidToC, a unified framework for studying video token compression across compression methods, backbone architectures, and time encoding strategies.
Through controlled comparisons across complementary video understanding tasks and training settings, we characterize the interactions among these design choices and highlight the importance of considering them jointly.
Our study provides empirical guidance for selecting, adapting, and evaluating compression strategies, supporting the development of efficient video MLLMs.

\section*{Limitations}

VidToC is a controlled empirical study rather than an exhaustive benchmark. Our architecture analysis compares one representative pure Transformer backbone and one representative hybrid Mamba-Transformer backbone at the 7B and 9B scales. The observed differences may indicate architecture-dependent trends, but the same quantitative trends may not hold for all Transformer or hybrid backbones.

We focus on portable compression strategies that can be inserted into existing video MLLMs. We do not exhaustively cover learnable query-based modules or architecture-specific compressors. Adaptive methods are excluded because their input-dependent token budgets are not directly comparable under our fixed-budget protocol. The multi-stage configurations are representative examples used to evaluate whether distributing token reduction across several stages can improve performance. We do not perform an exhaustive search over compression layers or method combinations.

Our default $16\times$ setting is an aggressive stress test. Other compression rates may change the ranking among methods, and lower compression rates may be preferable when memory and latency constraints are looser.

\section*{Acknowledgements}
Artificial intelligence was used to assist with drafting and editing manuscript text and with developing and debugging experimental code. The authors reviewed the resulting text and code and take full responsibility for the methods, results, and claims in this paper.

\bibliography{main}

\clearpage
\appendix
\renewcommand{\textfraction}{0.0}
\renewcommand{\topfraction}{1}
\renewcommand{\bottomfraction}{1}
\renewcommand{\floatpagefraction}{0.7}
\section{Implementation Details}
\label{app:implementation}

\subsection{Training and Evaluation Details}
\begin{table*}[tbp]
\centering
\small
\setlength{\tabcolsep}{3pt}
\renewcommand{\arraystretch}{1.2}
\begin{adjustbox}{max width=\textwidth}
\begin{tabular}{@{}ll>{\raggedright\arraybackslash}p{0.55\textwidth}@{}}
\toprule
\textbf{Stage} & \textbf{Data Type} & \textbf{Datasets} \\
\midrule
Pretraining & Image caption (3M) & CC12M~\cite{changpinyo2021cc12m} with PixelProse captions~\cite{singla2024pixelprose} \\
\midrule
\multirow{4}{*}{Instruction tuning} & Image instruction (2.8M) & LLaVA-OneVision~\cite{lillavaov} \\
\cline{2-3}
& Video instruction (1.3M) & LLaVA-Video~\cite{zhang2025llavavideo} \\
\cline{2-3}
& Detailed captioning (26K) & ActivityNet~\cite{caba2015activitynet}, COIN~\cite{tang2019coin}, HiREST~\cite{zala2023hirest}, ViTT~\cite{huang2020vitt}, YouCook2~\cite{zhou2018youcook2} \\
\cline{2-3}
& Temporal grounding (250K) & YT-Temporal~\cite{yang2023vid2seq}, DiDeMo~\cite{anne2017didemo}, QuerYD~\cite{oncescu2021queryd}, InternVid~\cite{wang2024internvid}, HowTo100M~\cite{miech2019howto100m}, annotated by VTG-IT~\cite{guo2025vtgllm}, TimeIT~\cite{ren2024timechat}, TimePro~\cite{zeng2025timesuite}, HTStep~\cite{afouras2023htstep}, and LongVid~\cite{li2024videochatflash} \\
\bottomrule
\end{tabular}
\end{adjustbox}
\caption{Training data recipe.}
\label{tab:training_data}
\end{table*}

This section provides additional details that complement the main paper.
The vision encoder resolution is 224$\times$224 and the patch size is 14, resulting in 256 visual tokens per frame before compression.
We feed features from the second-to-last layer of the vision encoder to the projector.
As SigLIP does not include a \texttt{[CLS]} token, we approximate \texttt{[CLS]} attention by averaging attention weights across all tokens, following VisionZip~\cite{yang2025visionzip}.
We train all models in two stages using the AdamW optimizer with a learning rate of $1e^{-5}$, cosine decay scheduling, a warmup ratio of 3\%, and a weight decay of 0.01.
The global batch size is 128 for pretraining and 64 for instruction tuning.
Both stages run for one epoch with BF16, Flash Attention 2, and DeepSpeed ZeRO Stage 1.
Table~\ref{tab:training_data} summarizes the datasets used in each training stage.

We use the following prompts for evaluation:
\begin{itemize}
  \raggedright
  \item \textbf{VideoMME}: ``\texttt{\{QUESTION\} Please provide your answer by stating the letter followed by the full option.}''
  \item \textbf{Charades}: ``\texttt{The video lasts for \{SECOND\} seconds, and \{FRAME\} frames are uniformly sampled from it. Investigate the video to identify the segment that matches the textual query \{QUERY\}, and output the start and end timestamps in seconds. The output format of the predicted timestamp should be like: 'start to end' seconds.}''
  \item \textbf{VDC}: Following the official evaluation script, the prompt is randomly selected from multiple candidates.
\end{itemize}

Unless otherwise noted, all training and evaluation runs use a global seed of 42, including stochastic GDR sampling and VDC prompt selection.
Each fine-tuned result comes from one training run, and each reported task score comes from one evaluation pass. All training and benchmark evaluation are conducted on eight NVIDIA H200 GPUs. Pretraining the uncompressed Qwen and Nano baselines take approximately 6 and 7.5 hours, respectively. Fine-tuning each uncompressed baseline takes approximately two days, and evaluating one model on all three benchmarks takes approximately two hours. Fine-tuning varies across compression methods. For reference, Spatial Pooling takes approximately two days.

\subsection{Compression Method Details}
\label{app:method_details}

Below we provide detailed implementation notes and parameter choices for some methods.

\begin{itemize}
  \item \textbf{Window Prune}: Uses a kernel size of $(4, 4)$. Unlike Spatial Pool, the center token within each grid cell is retained rather than pooled.
  \item \textbf{Spatial Pool}: Uses a kernel size of $(4, 4)$.
  \item \textbf{Spatiotemporal Pool}: Uses a kernel size of $(2, 4, 2)$ along the temporal, height, and width dimensions, respectively.
  \item \textbf{Pixel Unshuffle}: This operation changes the projector input dimension so it cannot be applied in a training-free manner. Unlike the other compression methods, which are trained only during the instruction tuning stage, Pixel Unshuffle must be included starting from the pretraining stage.
  \item \textbf{Temporal Prune}: All tokens in the first frame are retained. Tokens in subsequent frames are compared to their positionally corresponding tokens in the previous frame via cosine similarity, and the globally least similar tokens are retained.
  \item \textbf{GDR}: Greedy Duplicate Remove. We randomly sample $2\times\text{frame\_num}$ pivot tokens. The remaining tokens form a candidate pool. We evenly allocate the remaining token budget to each pivot. For each pivot, we retain the candidates with the lowest cosine similarity to that pivot and then remove the retained tokens from the candidate pool.
  \item \textbf{FPS}: Farthest Point Sampling. It iteratively selects tokens that maximize the minimum distance to already selected tokens.
  \item \textbf{Text-Attn}: For compression between layers $L$ and $L+1$, we use the query and key projections from layer $L+1$ to compute text-to-visual attention scores without completing that layer's forward pass. We then prune the visual tokens and run layer $L+1$ on the compressed sequence. At the projector level, we similarly use the query and key projections from the first LLM layer to compute the scores, prune the projected visual tokens, and run the LLM on the compressed sequence.
  \item \textbf{CLS-Sim}: Feature similarity between the \texttt{[CLS]} token and each visual token is computed as a scaled dot product followed by softmax normalization.
  \item \textbf{Bipartite Prune}: Discards the most similar token pairs based on bipartite matching. At most 24 token pairs are discarded per round.
  \item \textbf{Bipartite Merge}: Merges the most similar token pairs based on bipartite matching. At most 24 token pairs are merged per round.
  \item \textbf{Selective Slowfast}: Frames with high text-visual similarity are represented with $10 \times 10$ tokens, while frames with low text-visual similarity are downsampled to $2 \times 2$ tokens.
\end{itemize}

\subsection{Matched Token Budget}\label{app:matched_budget}

Using the budget relation in Section~\ref{subsec:compression_methods}, retaining a nonzero number of visual tokens requires $l/L < r$.
At the default retention rate $r=1/16$, this limits compression to positions with $l < L/16$. For the 28-layer Qwen2.5-7B, only layer~1 satisfies this constraint.
Nano has 56 computational blocks and uses L2, after the first sequence-mixing block and its following MLP.
The budget ratios are $1/28$ for both models,giving $r'=1/36$ at the selected position.

Matching the token budget does not guarantee equal FLOPs, latency, or peak memory.
For a Transformer, processing the full sequence in an early layer increases the quadratic attention cost even when the total token count across layers is fixed. Matching FLOPs would therefore require a even lower final retention rate.

\subsection{Retention Rate for Two-stage Compression}
We match the total number of visual tokens processed across decoder layers to a projector-level baseline with a $1/16$ retention rate.
The first method retains $1/4$ of the visual tokens at the projector, and the second compresses them after the second decoder layer.
For Qwen's 28 decoder layers, let $q$ be the second method's retention rate relative to its input.
Budget matching gives $2N/4+26Nq/4=28N/16$, where $N$ is the original visual-token count, and therefore $q=5/26\approx0.192$.

\subsection{Time Encoding Setting}
Qwen's MRoPE-I variant and Nano's RoPE and MRoPE-I variants use separately pretrained uncompressed reference models. Text timestamps are absent from pretraining. They are added only at evaluation in the training-free setting and from instruction tuning onward in the fine-tuned setting. Likewise, preserve and reset are selected only at evaluation for training-free results, but are used during instruction tuning for fine-tuned results.

For merging methods with \emph{preserve} strategy, we assign each output token an original position index. Spatial Pool and Selective SlowFast use the index of the central token in the corresponding spatial region. Spatiotemporal Pool uses the index of the central patch from the first frame of each temporal window. Bipartite Merge uses the earliest original index among the merged tokens.

\begin{figure*}[tp]
  \centering
  \includegraphics[width=0.8\linewidth]{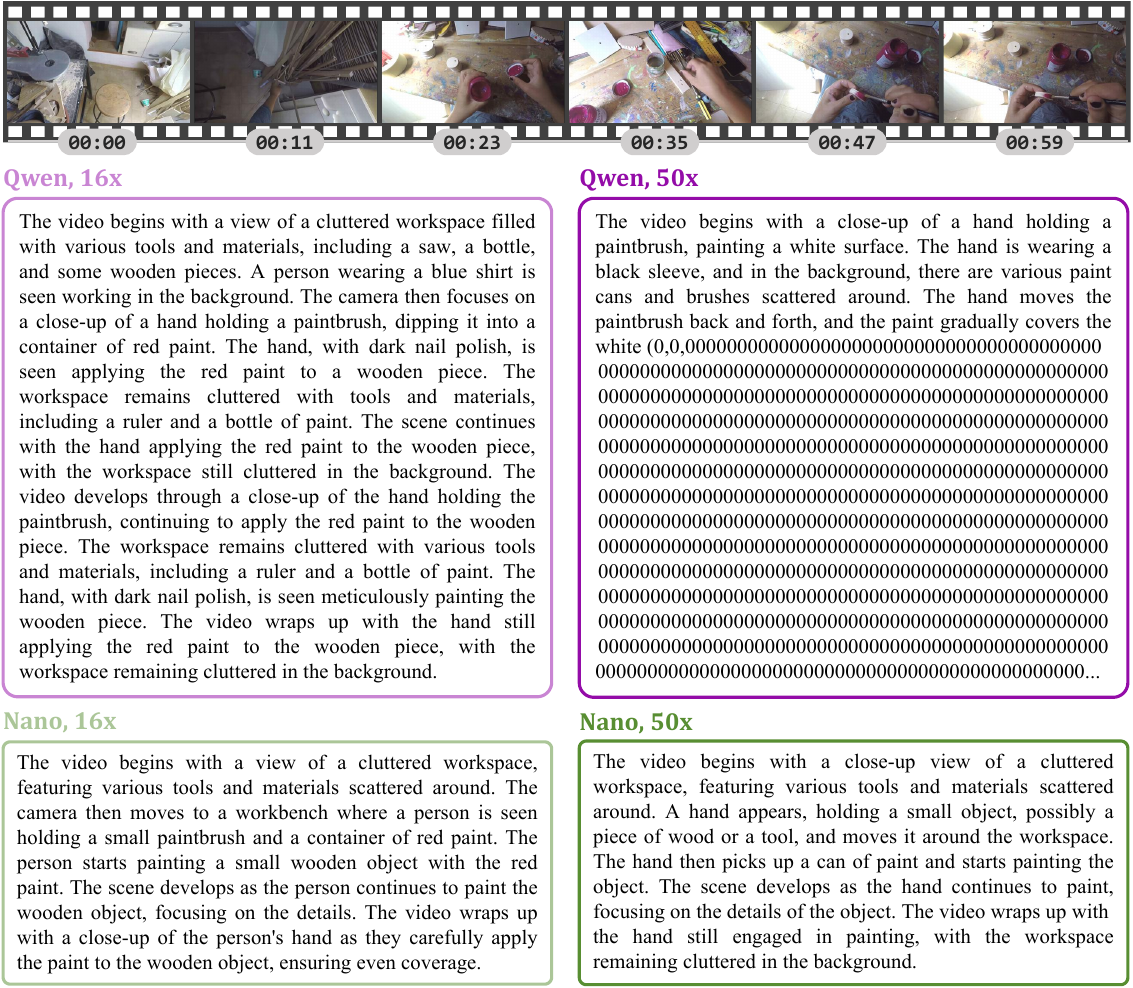}
  \caption{A qualitative case under Bipartite Merge with $50\times$ compression.}
  \label{fig:corrupted_case}
\end{figure*}

\section{Detailed Experimental Results}

\subsection{Main Results}
\label{app:main_results}

\begin{figure*}[tbp]
\centering
\includegraphics[width=1\linewidth]{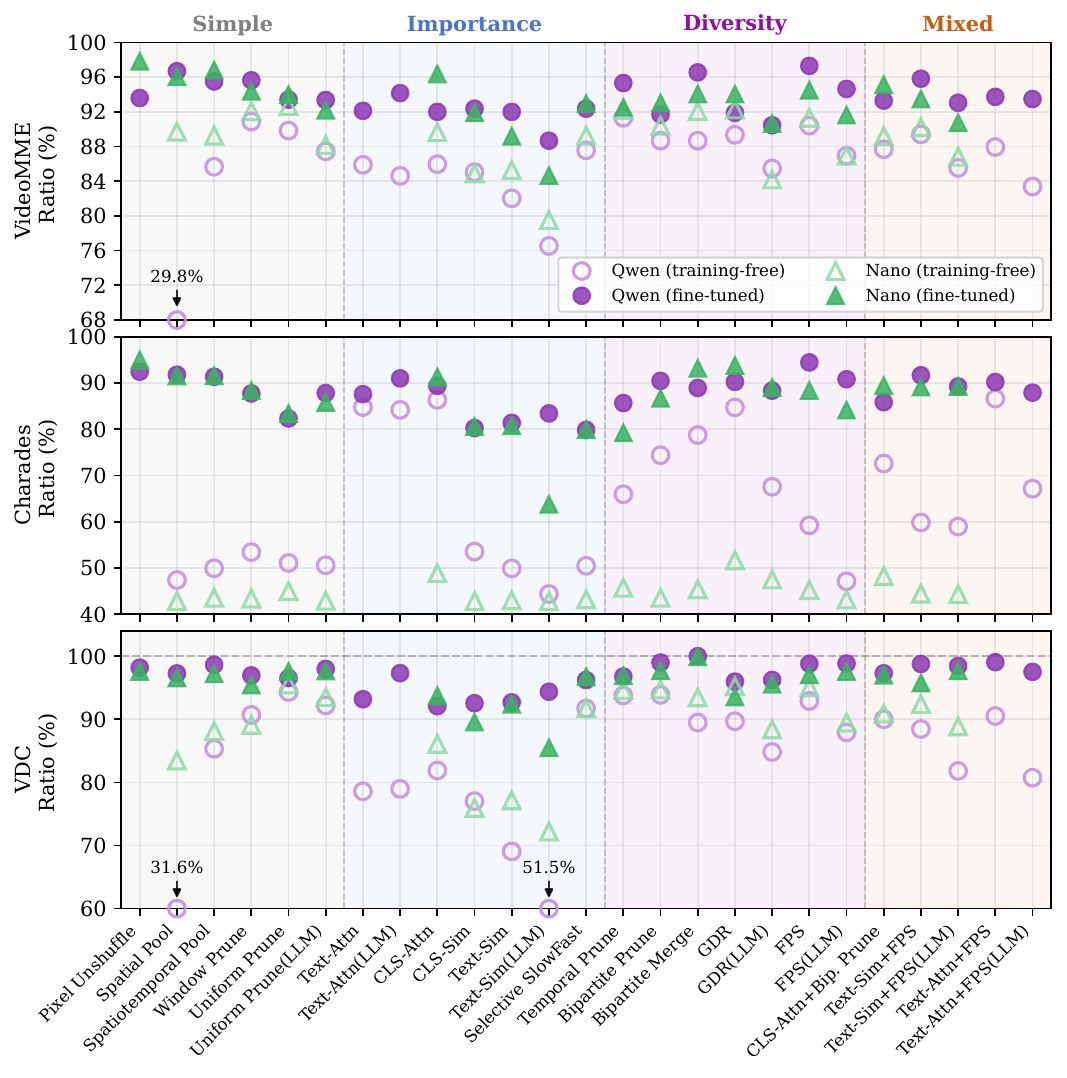}
\caption{Relative performance of compression methods on VideoMME, Charades, and VDC. Scores are normalized by the uncompressed reference of the corresponding backbone and task. The label LLM denotes compression at L1 on Qwen and L2 on Nano. Methods without this label apply compression at the projector.}
\label{fig:method_trends_app}
\end{figure*}

Tables~\ref{tab:results_qwen_tf}--\ref{tab:results_nano_ft} provide complete metrics for each compression method on Qwen and Nano.
Figure~\ref{fig:method_trends_app} visualizes relative performance on VideoMME, Charades, and VDC.

\subsection{Architecture Output Stability}
\label{app:output_stability}

To check whether output corruption is tied to one run or to Spatial Pool alone, we add two stability checks: Spatial Pool with another training seed (43) and Bipartite Merge at $50\times$ compression.
Table~\ref{tab:output_stability} reports task scores and corruption rates (the proportion of samples whose generated responses contain corrupted text), and Figure~\ref{fig:corrupted_case} shows a VDC case under Bipartite Merge.

The seed-43 Spatial Pool run is less severe than seed 42, but it keeps the same pattern: Qwen still corrupts many outputs on VideoMME and VDC, while Charades is much less affected.
Thus, the Spatial Pool failure is not a single-seed artifact, although its severity varies across runs.
Bipartite Merge at $50\times$ gives a useful contrast.
Even under a more aggressive compression, it produces far fewer corrupted outputs than Spatial Pool.
This is because Bipartite Merge combines visually similar tokens, so each merged token stays close to the original local representation.

\begin{table}[tp]
\centering
\small
\setlength{\tabcolsep}{3pt}
\begin{adjustbox}{max width=\columnwidth}
\begin{tabular}{@{}l|c|cc|cc|cc@{}}
\toprule
\multirow{2}{*}{Method / setting} & \multirow{2}{*}{Backbone} & \multicolumn{2}{c|}{VideoMME} & \multicolumn{2}{c|}{Charades} & \multicolumn{2}{c}{VDC} \\
& & Acc. & CR & mIoU & CR & Acc. & CR \\
\midrule
\textcolor{gray}{Spatial Pool, seed 42} & \textcolor{gray}{Qwen} & \textcolor{gray}{17.96} & \textcolor{gray}{69.85\%} & \textcolor{gray}{16.04} & \textcolor{gray}{8.18\%} & \textcolor{gray}{14.12} & \textcolor{gray}{81.30\%} \\
Spatial Pool, seed 43 & Qwen & 33.04 & 32.04\% & 16.32 & 1.93\% & 23.81 & 40.70\% \\
\midrule
Bipartite Merge, $50\times$ & Qwen & 48.56 & 0.26\% & 16.91 & 0.00\% & 34.27 & 1.07\% \\
Bipartite Merge, $50\times$ & Nano & 49.04 & 0.00\% & 13.81 & 0.00\% & 35.66 & 0.00\% \\
\bottomrule
\end{tabular}
\end{adjustbox}
\caption{Output stability under different settings. CR: Corruption Rate, the proportion of samples containing corrupted text. The gray row repeats the result from the main table for reference.}
\label{tab:output_stability}
\end{table}

\begin{figure*}[tp]
\centering
\includegraphics[width=0.8\linewidth]{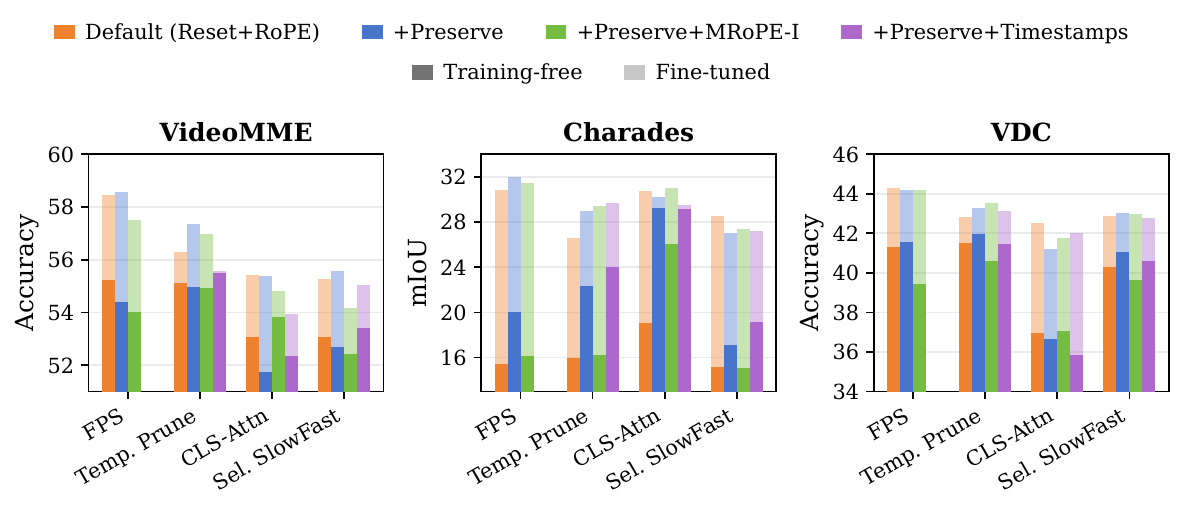}
\vspace{2pt}
\includegraphics[width=0.8\linewidth]{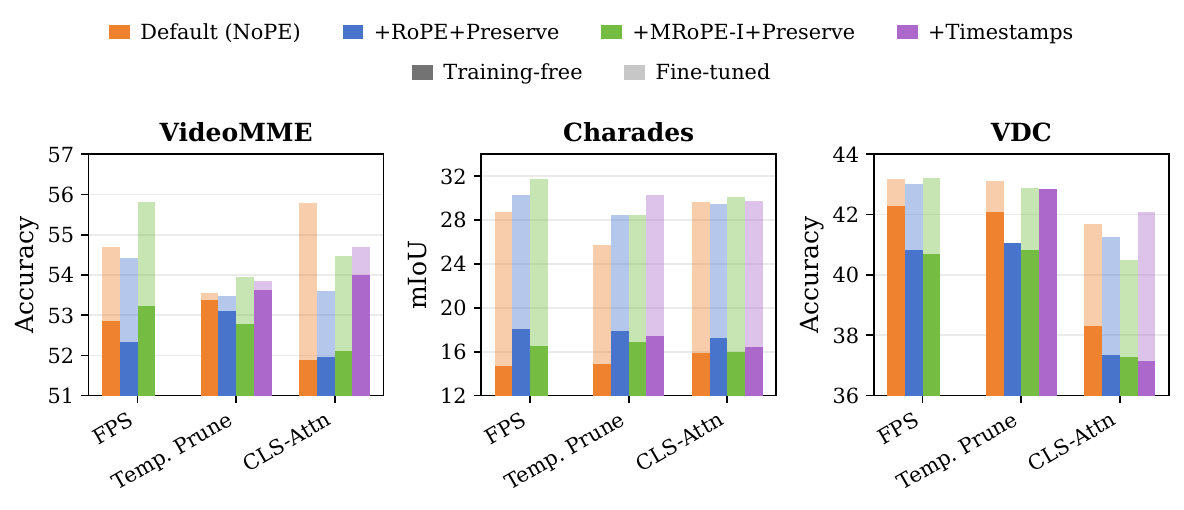}
\caption{Full time-encoding results on Qwen2.5-7B and Nano-9B-v2. NoPE means no position encoding.}
\label{fig:time_encoding_full_app}
\end{figure*}

\subsection{Projector-Level vs. LLM-Internal Compression at $8\times$}
\label{app:position_8x}

Table~\ref{tab:compression_position_8x} compares projector-level and LLM-internal compression at a milder $8\times$ compression rate to complement the main paper's $16\times$ comparison.
For each method, we identify the best training-free LLM-internal position and fine-tune only that configuration.
Training-free performance generally declines as compression moves deeper, making L1 the selected position for both FPS and Text-Attn.
After fine-tuning, FPS-L1 remains slightly below projector-level compression, whereas Text-Attn-L1 exceeds its projector-level counterpart on all three tasks.

\begin{table}[tp]
\centering
\small
\setlength{\tabcolsep}{3pt}
\begin{adjustbox}{max width=\columnwidth}
\begin{tabular}{@{}l|c|cc|cc|cc@{}}
\toprule
\multirow{2}{*}{Method} & \multirow{2}{*}{Pos.} & \multicolumn{2}{c|}{VideoMME} & \multicolumn{2}{c|}{Charades} & \multicolumn{2}{c}{VDC} \\
& & TF & FT & TF & FT & TF & FT \\
\midrule
FPS & Proj & 56.19 & 59.78 & 24.59 & 32.92 & 43.98 & 44.37 \\
FPS & L1 & 56.00 & 59.07 & 22.54 & 32.05 & 42.81 & 44.26 \\
FPS & L2 & 54.63 & -- & 20.10 & -- & 41.63 & -- \\
FPS & L3 & 52.70 & -- & 15.73 & -- & 38.13 & -- \\
\midrule
Text-Attn & Proj & 54.19 & 56.70 & 30.36 & 29.88 & 38.81 & 42.75 \\
Text-Attn & L1 & 53.11 & 57.41 & 30.53 & 30.60 & 37.50 & 43.36 \\
Text-Attn & L2 & 52.67 & -- & 28.12 & -- & 35.99 & -- \\
Text-Attn & L3 & 44.07 & -- & 17.19 & -- & 20.86 & -- \\
\bottomrule
\end{tabular}
\end{adjustbox}
\caption{
  Comparison between projector-level and LLM-internal compression at 8$\times$ on Qwen2.5-7B.
  Only the best training-free LLM-internal position reports fine-tuned results.
}
\label{tab:compression_position_8x}
\end{table}

\subsection{Time Encoding Results}
\label{app:time_encoding}

Figure~\ref{fig:time_encoding_full_app} reports the time encoding results for Qwen and Nano across VideoMME, Charades, and VDC.
On Nano, adding RoPE+Preserve also improves training-free performance on Charades, but the gains are more modest.
After fine-tuning, the benefit remains visible for methods with variable token counts per frame, matching the pattern observed on Qwen.

\begin{figure*}[tp]
  \centering
  \includegraphics[width=0.9\linewidth]{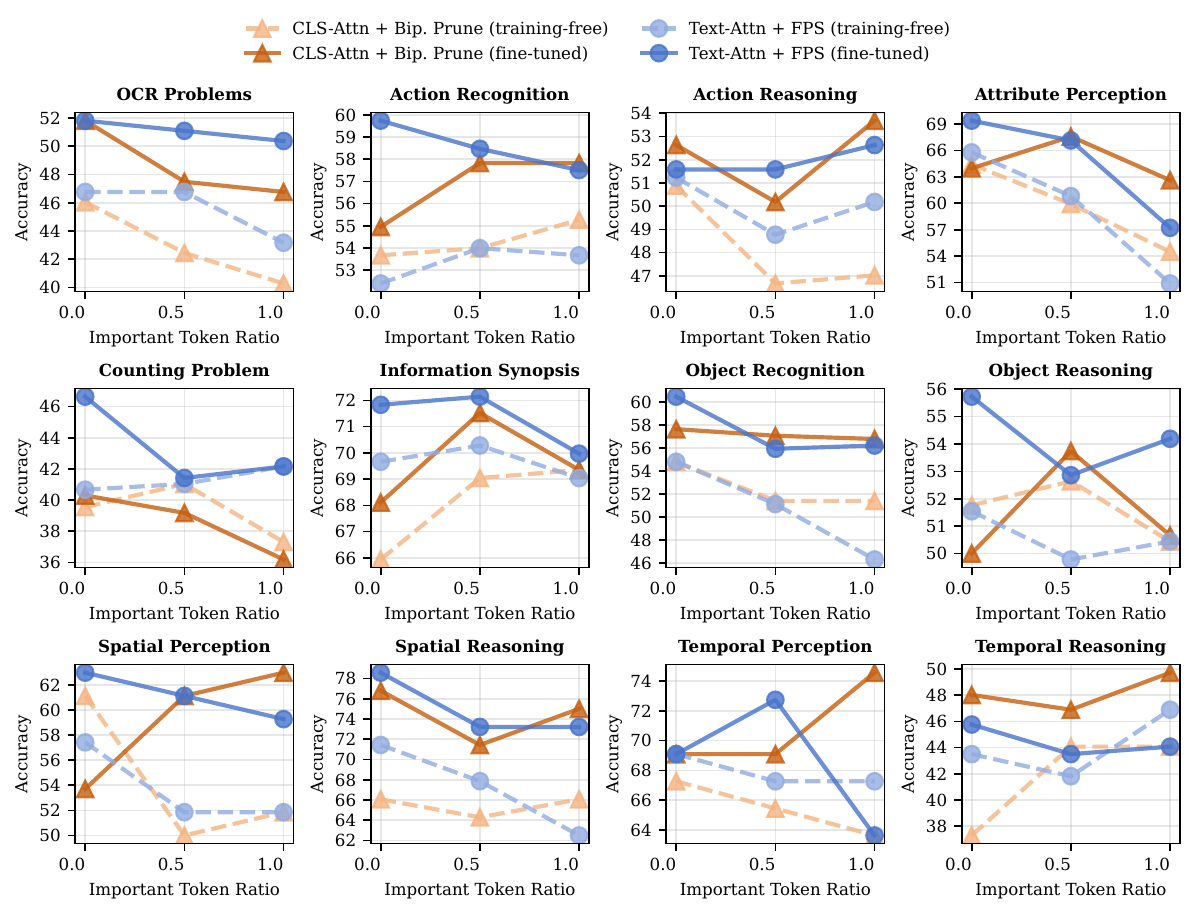}
  \caption{VideoMME subtask preference.}
  \label{fig:videomme_subtask}
\end{figure*}

\subsection{Importance-Diversity Preference of VideoMME Subtasks}

To understand the mixed VideoMME results reported in the main paper, we examine its 12 subtasks across different importance ratios. In the training-free setting, OCR Problems, Attribute Perception, Object Recognition, and Spatial Perception decline as the importance ratio increases. These tasks often require broad visual scanning, so stricter importance sampling may favor foreground content while discarding peripheral details.

Action Recognition and Temporal Reasoning remain stable or improve slightly at higher importance ratios, where focusing on salient regions may suppress static background noise. Information Synopsis and Action Reasoning vary non-monotonically, consistent with their need for both global context and local actions. Since most subtasks benefit more from broad visual coverage, the overall VideoMME score declines as the importance ratio rises.

\section{Efficiency Analysis}
\label{app:efficiency}

Tables~\ref{tab:efficiency_qwen} and~\ref{tab:efficiency_nano} report efficiency on one NVIDIA M402 GPU using the same 20 video inputs at 32 and 256 frames, with two warm-up samples and fixed 128-token generation.
Prefill includes the vision encoder, projector, and LLM, excluding preprocessing and data transfer.
Decoding throughput covers the subsequent 127 tokens.
Peak allocated memory is reported in GiB.
Compression time measures cumulative compressor execution across prefill and decoding.
Projector-level compression uses a retention rate of 1/16, and Qwen L1 and Nano L2 use 1/36 according to the matched token budget.

At 256 frames, Qwen L1 increases prefill latency by 18.4\% to 23.8\% for 10 of 14 methods.
Nano shows smaller differences, with L2 increasing latency by 1.0\% to 4.8\% for 8 of 12 methods.
Peak memory increases for every internal counterpart on both backbones.
Together with the training-free performance results, these measurements favor projector-level compression on both performance and efficiency for most evaluated methods.
If LLM-internal compression retains more tokens to improve performance, its computational cost increases further.

FPS and GDR are exceptions to this latency pattern because their selection algorithms are iterative.
For $N$ input tokens, $K$ retained tokens, and feature dimension $d$, FPS requires $K$ sequential rounds and our GDR implementation uses $\lfloor K/8\rfloor$ pivot rounds, giving both $O(NKd)$ similarity cost. 
The lower internal retention rate reduces $K$ and thus the number of rounds. This also explains why FPS and GDR have substantially higher prefill latency and are even slower than the uncompressed baseline at 256 frames when applied at the projector.
GDR and FPS-based combinations show similar behavior.
Their speedups come from reduced selection overhead rather than cheaper LLM processing.

\begin{table*}[p]
\centering
\small
\setlength{\tabcolsep}{1pt}
\begin{tabularx}{0.8\textwidth}{
  @{}
  >{\raggedright\arraybackslash}p{0.17\textwidth} |
  *{3}{>{\centering\arraybackslash}X} |
  *{3}{>{\centering\arraybackslash}X}
  @{}
}
\toprule
\multirow{2}{*}{Method} &
\multicolumn{3}{c|}{Qwen2.5-7B} &
\multicolumn{3}{c}{Nano-9B-v2} \\
\cmidrule(l{1pt}r{1pt}){2-4}
\cmidrule(l{1pt}r{1pt}){5-7}
& VideoMME & Charades & VDC
& VideoMME & Charades & VDC \\
\midrule
Uncompressed
& 58.48 & 32.13 & 45.61
& 34.70 & 31.28 & 44.65 \\
\midrule
\rowcolor{gray!10}
\multicolumn{7}{@{}l}{\emph{Simple methods}} \\
Uniform Prune
& \dropscore{1}{54.70} & \dropscore{4}{17.18} & \dropscore{1}{42.45}
& 55.59 & \dropscore{4}{14.58} & 42.83 \\
Window Prune
& 56.30 & \dropscore{4}{18.11} & \dropscore{2}{40.32}
& 57.26 & \dropscore{4}{14.13} & \dropscore{2}{40.18} \\
Spatial Pool
& \dropscore{4}{20.52} & \dropscore{4}{16.12} & \dropscore{4}{16.33}
& 54.30 & \dropscore{4}{14.08} & \dropscore{3}{32.10} \\
Spatiotemporal Pool
& \dropscore{2}{52.56} & \dropscore{4}{16.87} & \dropscore{2}{38.62}
& 55.74 & \dropscore{4}{14.17} & \dropscore{2}{39.30} \\
\midrule
\rowcolor{gray!10}
\multicolumn{7}{@{}l}{\emph{Atomic methods}} \\
CLS-Attn
& \dropscore{1}{52.85} & \dropscore{2}{28.86} & \dropscore{2}{36.72}
& 54.00 & \dropscore{4}{16.09} & \dropscore{2}{38.16} \\
CLS-Sim
& \dropscore{2}{51.93} & \dropscore{4}{18.18} & \dropscore{3}{34.47}
& 52.56 & \dropscore{4}{14.00} & \dropscore{3}{33.98} \\
Bipartite Prune
& \dropscore{1}{54.41} & \dropscore{3}{24.67} & \dropscore{1}{42.55}
& 56.30 & \dropscore{4}{14.20} & 42.52 \\
Bipartite Merge
& \dropscore{1}{54.15} & \dropscore{2}{25.97} & \dropscore{2}{40.11}
& 55.89 & \dropscore{4}{14.98} & \dropscore{1}{41.32} \\
Temporal Prune
& 58.74 & \dropscore{3}{22.57} & \dropscore{1}{42.70}
& 57.89 & \dropscore{4}{15.02} & 42.54 \\
Selective Slowfast
& \dropscore{1}{54.11} & \dropscore{4}{17.19} & \dropscore{1}{41.10}
& 56.00 & \dropscore{4}{14.09} & \dropscore{1}{41.40} \\
GDR
& \dropscore{1}{54.44} & \dropscore{2}{27.49} & \dropscore{2}{40.20}
& 54.96 & \dropscore{4}{17.02} & \dropscore{1}{41.21} \\
FPS
& 56.33 & \dropscore{4}{19.79} & \dropscore{1}{42.15}
& 55.78 & \dropscore{4}{14.69} & \dropscore{1}{42.36} \\
Text-Attn
& \dropscore{2}{52.00} & \dropscore{2}{28.08} & \dropscore{3}{35.39}
& -- & -- & -- \\
Text-Sim
& \dropscore{2}{49.74} & \dropscore{4}{17.05} & \dropscore{4}{31.48}
& 51.37 & \dropscore{4}{13.99} & \dropscore{3}{34.36} \\
\midrule
\rowcolor{gray!10}
\multicolumn{7}{@{}l}{\emph{Mixed methods}} \\
CLS-Attn+Bip. Prune
& \dropscore{1}{52.89} & \dropscore{3}{24.35} & \dropscore{2}{40.16}
& 54.44 & \dropscore{4}{15.77} & \dropscore{1}{41.20} \\
Text-Sim+FPS
& \dropscore{1}{55.04} & \dropscore{4}{20.07} & \dropscore{2}{40.55}
& 55.30 & \dropscore{4}{14.39} & \dropscore{1}{41.75} \\
Text-Attn+FPS
& \dropscore{1}{53.93} & \dropscore{2}{28.57} & \dropscore{2}{40.31}
& -- & -- & -- \\
\bottomrule
\end{tabularx}
\caption{Training-free results with inputs of up to 256 frames.}
\label{tab:results_256_frames}
\end{table*}

\section{Scaling Evaluation to 256 Frames}

The main experiments train and evaluate models with at most 32 frames.
Here, we further evaluate the training-free setting with a maximum of 256 input frames without retraining.
The effect of using 256 frames is most visible on VideoMME, as Charades and VDC contain relatively short videos, for which 32 frames already cover most relevant visual content.
The uncompressed baseline drops on VideoMME as the input substantially exceeds the training context length.

Overall, the ranking pattern on Qwen remains largely stable.
Temporal Prune remains the strongest method on VideoMME, and the top-ranking methods on Charades and VDC mostly stay within the same group.
Nano shows larger rank shifts on VideoMME, where Temporal Prune, Window Prune, and Bipartite Prune become stronger.

\section{License}

Our use of third-party datasets is solely for academic research and does not involve commercial use. The materials used in this work include datasets or annotations with non-commercial or academic-only restrictions, including VideoMME, Charades, COIN annotations, HT-Step annotations, InternVid, TimeLens-Bench, and LLaVA-Video-178K. We comply with the applicable terms of each source.

\begin{table*}[tp]
\centering
\small
\setlength{\tabcolsep}{1.6pt}
\begin{adjustbox}{max width=\textwidth}
\begin{tabularx}{\textwidth}{@{}l@{\hspace{8pt}}|*{4}{>{\centering\arraybackslash}X}|*{4}{>{\centering\arraybackslash}X}|*{2}{>{\centering\arraybackslash}X}@{}}
\toprule
\multirow{2}{*}{Method} & \multicolumn{4}{c|}{VideoMME} & \multicolumn{4}{c|}{Charades} & \multicolumn{2}{c}{VDC} \\
& Avg. & Short & Medium & Long & mIoU & R@.3 & R@.5 & R@.7 & Acc. & Score \\
\midrule
Uncompressed & 60.19 & 72.44 & 58.22 & 49.89 & 33.83 & 51.38 & 24.89 & 11.03 & 44.74 & 2.32 \\
\midrule
\rowcolor{gray!10}
\multicolumn{11}{@{}l}{\emph{Simple methods}} \\
Uniform Prune & \dropscore{2}{54.07} & \dropscore{2}{62.11} & \dropscore{1}{52.44} & 47.67 & \dropscore{4}{17.29} & \dropscore{4}{25.60} & \dropscore{4}{11.54} & \dropscore{4}{5.38} & \dropscore{1}{42.20} & \dropscore{1}{2.18} \\
Window Prune & \dropscore{1}{54.70} & \dropscore{2}{63.78} & \dropscore{2}{51.67} & 48.67 & \dropscore{4}{18.08} & \dropscore{4}{27.00} & \dropscore{4}{12.10} & \dropscore{4}{5.65} & \dropscore{1}{40.58} & \dropscore{1}{2.11} \\
Spatial Pool & \dropscore{4}{17.96} & \dropscore{4}{17.56} & \dropscore{4}{17.33} & \dropscore{4}{19.00} & \dropscore{4}{16.04} & \dropscore{4}{23.76} & \dropscore{4}{11.51} & \dropscore{4}{5.29} & \dropscore{4}{14.12} & \dropscore{4}{0.75} \\
Spatiotemporal Pool & \dropscore{2}{51.56} & \dropscore{2}{59.22} & \dropscore{2}{49.78} & \dropscore{1}{45.67} & \dropscore{4}{16.90} & \dropscore{4}{25.01} & \dropscore{4}{11.78} & \dropscore{4}{5.56} & \dropscore{2}{38.18} & \dropscore{2}{2.01} \\
Pixel Unshuffle & -- & -- & -- & -- & -- & -- & -- & -- & -- & -- \\
\midrule
\rowcolor{gray!10}
\multicolumn{11}{@{}l}{\emph{Atomic methods}} \\
CLS-Attn & \dropscore{2}{51.74} & \dropscore{2}{59.11} & \dropscore{2}{50.78} & \dropscore{1}{45.33} & \dropscore{2}{29.21} & \dropscore{2}{43.62} & \dropscore{2}{20.96} & \dropscore{2}{9.81} & \dropscore{2}{36.64} & \dropscore{2}{1.92} \\
CLS-Sim & \dropscore{2}{51.19} & \dropscore{3}{56.78} & \dropscore{2}{50.22} & \dropscore{1}{46.56} & \dropscore{4}{18.12} & \dropscore{4}{26.70} & \dropscore{4}{12.64} & \dropscore{4}{5.89} & \dropscore{3}{34.47} & \dropscore{3}{1.80} \\
Bipartite Prune & \dropscore{2}{53.37} & \dropscore{2}{63.22} & \dropscore{2}{51.89} & \dropscore{1}{45.00} & \dropscore{3}{25.16} & \dropscore{3}{38.18} & \dropscore{3}{17.60} & \dropscore{4}{7.58} & \dropscore{1}{42.00} & \dropscore{1}{2.18} \\
Bipartite Merge & \dropscore{2}{53.37} & \dropscore{2}{62.33} & \dropscore{2}{51.89} & \dropscore{1}{45.89} & \dropscore{3}{26.64} & \dropscore{3}{40.32} & \dropscore{3}{19.21} & \dropscore{3}{8.71} & \dropscore{2}{40.04} & \dropscore{2}{2.08} \\
Temporal Prune & \dropscore{1}{54.96} & \dropscore{2}{62.67} & \dropscore{1}{54.00} & 48.22 & \dropscore{4}{22.31} & \dropscore{4}{33.51} & \dropscore{4}{15.19} & \dropscore{4}{6.87} & \dropscore{1}{41.96} & \dropscore{1}{2.18} \\
Selective Slowfast & \dropscore{2}{52.70} & \dropscore{2}{59.78} & \dropscore{2}{51.33} & \dropscore{1}{47.00} & \dropscore{4}{17.08} & \dropscore{4}{25.39} & \dropscore{4}{11.95} & \dropscore{4}{5.53} & \dropscore{1}{41.04} & \dropscore{1}{2.12} \\
GDR & \dropscore{2}{53.78} & \dropscore{2}{63.67} & \dropscore{2}{51.67} & \dropscore{1}{46.00} & \dropscore{2}{28.66} & \dropscore{2}{43.12} & \dropscore{2}{20.87} & \dropscore{2}{9.37} & \dropscore{2}{40.13} & \dropscore{2}{2.08} \\
FPS & \dropscore{1}{54.41} & \dropscore{2}{63.78} & \dropscore{2}{51.11} & 48.33 & \dropscore{4}{20.04} & \dropscore{4}{29.94} & \dropscore{4}{13.50} & \dropscore{4}{5.92} & \dropscore{1}{41.57} & \dropscore{1}{2.17} \\
Text-Attn & \dropscore{2}{51.70} & \dropscore{3}{57.56} & \dropscore{2}{50.33} & \dropscore{1}{47.22} & \dropscore{2}{28.68} & \dropscore{2}{42.97} & \dropscore{2}{20.58} & \dropscore{2}{9.43} & \dropscore{3}{35.16} & \dropscore{2}{1.86} \\
Text-Sim & \dropscore{2}{49.37} & \dropscore{3}{54.22} & \dropscore{2}{48.67} & \dropscore{1}{45.22} & \dropscore{4}{16.89} & \dropscore{4}{25.16} & \dropscore{4}{11.48} & \dropscore{4}{5.29} & \dropscore{4}{30.90} & \dropscore{3}{1.63} \\
\midrule
\rowcolor{gray!10}
\multicolumn{11}{@{}l}{\emph{Mixed methods}} \\
CLS-Attn+Bip. Prune & \dropscore{2}{52.78} & \dropscore{2}{61.67} & \dropscore{2}{50.89} & \dropscore{1}{45.78} & \dropscore{3}{24.56} & \dropscore{3}{37.05} & \dropscore{4}{17.25} & \dropscore{3}{7.79} & \dropscore{1}{40.27} & \dropscore{1}{2.10} \\
Text-Sim+FPS & \dropscore{2}{53.81} & \dropscore{2}{63.67} & \dropscore{2}{51.00} & \dropscore{1}{46.78} & \dropscore{4}{20.25} & \dropscore{4}{30.24} & \dropscore{4}{13.47} & \dropscore{4}{6.16} & \dropscore{2}{39.58} & \dropscore{2}{2.08} \\
Text-Attn+FPS & \dropscore{2}{52.93} & \dropscore{2}{62.56} & \dropscore{2}{50.67} & \dropscore{1}{45.56} & \dropscore{2}{29.30} & \dropscore{2}{44.04} & \dropscore{2}{20.87} & \dropscore{2}{9.37} & \dropscore{1}{40.50} & \dropscore{1}{2.10} \\
\bottomrule
\end{tabularx}
\end{adjustbox}
\caption{Training-free results on Qwen2.5-7B.}
\label{tab:results_qwen_tf}
\end{table*}

\begin{table*}[tp]
\centering
\small
\setlength{\tabcolsep}{1.6pt}
\begin{adjustbox}{max width=\textwidth}
\begin{tabularx}{\textwidth}{@{}l@{\hspace{8pt}}|*{4}{>{\centering\arraybackslash}X}|*{4}{>{\centering\arraybackslash}X}|*{2}{>{\centering\arraybackslash}X}@{}}
\toprule
\multirow{2}{*}{Method} & \multicolumn{4}{c|}{VideoMME} & \multicolumn{4}{c|}{Charades} & \multicolumn{2}{c}{VDC} \\
& Avg. & Short & Medium & Long & mIoU & R@.3 & R@.5 & R@.7 & Acc. & Score \\
\midrule
Uncompressed & 60.19 & 72.44 & 58.22 & 49.89 & 33.83 & 51.38 & 24.89 & 11.03 & 44.74 & 2.32 \\
\midrule
\rowcolor{gray!10}
\multicolumn{11}{@{}l}{\emph{Simple methods}} \\
Uniform Prune & \dropscore{1}{56.22} & \dropscore{2}{64.00} & \dropscore{1}{55.22} & 49.44 & \dropscore{2}{27.85} & \dropscore{2}{41.81} & \dropscore{2}{20.13} & \dropscore{3}{8.68} & 43.19 & 2.23 \\
Window Prune & 57.56 & \dropscore{1}{68.44} & 55.33 & 48.89 & \dropscore{2}{29.68} & \dropscore{2}{44.34} & \dropscore{2}{21.38} & \dropscore{2}{9.84} & 43.38 & 2.25 \\
Spatial Pool & 58.19 & 69.00 & 55.89 & 49.67 & \dropscore{1}{31.05} & \dropscore{1}{47.04} & \dropscore{1}{22.45} & \dropscore{2}{9.84} & 43.52 & 2.25 \\
Spatiotemporal Pool & 57.48 & \dropscore{1}{68.56} & \dropscore{1}{54.33} & 49.56 & \dropscore{1}{30.90} & \dropscore{1}{46.54} & \dropscore{2}{21.94} & \dropscore{2}{9.78} & 44.13 & 2.29 \\
Pixel Unshuffle & \dropscore{1}{56.33} & \dropscore{1}{67.22} & \dropscore{1}{53.44} & 48.33 & \dropscore{1}{31.27} & \dropscore{1}{47.34} & \dropscore{1}{22.54} & \dropscore{1}{10.23} & 43.93 & 2.29 \\
\midrule
\rowcolor{gray!10}
\multicolumn{11}{@{}l}{\emph{Atomic methods}} \\
CLS-Attn & \dropscore{1}{55.37} & \dropscore{2}{63.44} & \dropscore{1}{53.89} & 48.78 & \dropscore{2}{30.24} & \dropscore{2}{45.70} & \dropscore{2}{21.91} & \dropscore{2}{9.69} & \dropscore{1}{41.20} & \dropscore{1}{2.16} \\
CLS-Sim & \dropscore{1}{55.59} & \dropscore{2}{63.22} & \dropscore{1}{54.56} & 49.00 & \dropscore{2}{27.15} & \dropscore{3}{40.86} & \dropscore{3}{19.18} & \dropscore{3}{8.39} & \dropscore{1}{41.41} & \dropscore{1}{2.16} \\
Bipartite Prune & \dropscore{1}{55.19} & \dropscore{2}{64.33} & \dropscore{1}{53.56} & 47.67 & \dropscore{1}{30.60} & \dropscore{1}{46.39} & \dropscore{2}{22.39} & \dropscore{2}{9.87} & 44.29 & 2.30 \\
Bipartite Merge & 58.11 & \dropscore{1}{68.00} & 57.33 & 49.00 & \dropscore{2}{30.08} & \dropscore{2}{45.29} & \dropscore{2}{21.38} & \dropscore{2}{9.52} & 44.74 & 2.32 \\
Temporal Prune & 57.37 & \dropscore{1}{66.22} & 55.78 & 50.11 & \dropscore{2}{28.98} & \dropscore{2}{43.86} & \dropscore{2}{21.41} & \dropscore{2}{9.01} & 43.30 & 2.24 \\
Selective Slowfast & \dropscore{1}{55.59} & \dropscore{2}{63.33} & \dropscore{1}{54.00} & 49.44 & \dropscore{3}{27.01} & \dropscore{3}{40.62} & \dropscore{3}{19.57} & \dropscore{3}{8.39} & 43.05 & 2.23 \\
GDR & \dropscore{1}{55.30} & \dropscore{2}{63.22} & \dropscore{1}{54.00} & 48.67 & \dropscore{1}{30.53} & \dropscore{1}{46.51} & \dropscore{2}{21.56} & \dropscore{2}{9.49} & 42.94 & 2.24 \\
FPS & 58.56 & 69.67 & 57.78 & 48.22 & \dropscore{1}{31.95} & \dropscore{1}{48.44} & \dropscore{1}{22.99} & \dropscore{1}{10.44} & 44.21 & 2.30 \\
Text-Attn & \dropscore{1}{55.44} & \dropscore{2}{62.56} & \dropscore{1}{54.56} & 49.22 & \dropscore{2}{29.63} & \dropscore{2}{45.05} & \dropscore{2}{21.17} & \dropscore{2}{9.16} & \dropscore{1}{41.69} & \dropscore{1}{2.16} \\
Text-Sim & \dropscore{1}{55.37} & \dropscore{2}{65.00} & \dropscore{2}{52.22} & 48.89 & \dropscore{2}{27.53} & \dropscore{2}{41.75} & \dropscore{3}{19.57} & \dropscore{3}{8.77} & \dropscore{1}{41.47} & \dropscore{1}{2.17} \\
\midrule
\rowcolor{gray!10}
\multicolumn{11}{@{}l}{\emph{Mixed methods}} \\
CLS-Attn+Bip. Prune & \dropscore{1}{56.15} & \dropscore{1}{66.22} & \dropscore{1}{54.22} & 48.00 & \dropscore{2}{29.04} & \dropscore{2}{43.77} & \dropscore{2}{20.81} & \dropscore{2}{9.43} & 43.53 & 2.27 \\
Text-Sim+FPS & 57.67 & 69.22 & 55.89 & 47.89 & \dropscore{1}{31.01} & \dropscore{1}{46.60} & \dropscore{2}{22.39} & \dropscore{1}{10.08} & 44.19 & 2.28 \\
Text-Attn+FPS & \dropscore{1}{56.41} & \dropscore{1}{66.22} & \dropscore{1}{53.67} & 49.33 & \dropscore{1}{30.52} & \dropscore{2}{45.94} & \dropscore{2}{22.09} & \dropscore{2}{9.55} & 44.32 & 2.30 \\
\bottomrule
\end{tabularx}
\end{adjustbox}
\caption{Fine-tuned results on Qwen2.5-7B.}
\label{tab:results_qwen_ft}
\end{table*}

\begin{table*}[tp]
\centering
\small
\setlength{\tabcolsep}{1.6pt}
\begin{adjustbox}{max width=\textwidth}
\begin{tabularx}{\textwidth}{@{}l@{\hspace{8pt}}|*{4}{>{\centering\arraybackslash}X}|*{4}{>{\centering\arraybackslash}X}|*{2}{>{\centering\arraybackslash}X}@{}}
\toprule
\multirow{2}{*}{Method} & \multicolumn{4}{c|}{VideoMME} & \multicolumn{4}{c|}{Charades} & \multicolumn{2}{c}{VDC} \\
& Avg. & Short & Medium & Long & mIoU & R@.3 & R@.5 & R@.7 & Acc. & Score \\
\midrule
Uncompressed & 57.89 & 68.78 & 56.44 & 48.44 & 32.50 & 48.74 & 23.55 & 10.14 & 44.48 & 2.32 \\
\midrule
\rowcolor{gray!10}
\multicolumn{11}{@{}l}{\emph{Simple methods}} \\
Uniform Prune & \dropscore{1}{53.63} & \dropscore{2}{61.33} & \dropscore{1}{52.78} & 46.78 & \dropscore{4}{14.62} & \dropscore{4}{21.91} & \dropscore{4}{10.56} & \dropscore{4}{4.85} & 42.48 & \dropscore{1}{2.20} \\
Window Prune & \dropscore{1}{53.30} & \dropscore{2}{59.56} & \dropscore{1}{52.33} & 48.00 & \dropscore{4}{14.10} & \dropscore{4}{20.87} & \dropscore{4}{10.08} & \dropscore{4}{4.64} & \dropscore{2}{39.64} & \dropscore{2}{2.07} \\
Spatial Pool & \dropscore{2}{51.93} & \dropscore{2}{58.67} & \dropscore{1}{50.89} & 46.22 & \dropscore{4}{13.93} & \dropscore{4}{20.58} & \dropscore{4}{10.08} & \dropscore{4}{4.76} & \dropscore{2}{37.13} & \dropscore{2}{1.95} \\
Spatiotemporal Pool & \dropscore{2}{51.67} & \dropscore{2}{57.44} & \dropscore{1}{51.22} & 46.33 & \dropscore{4}{14.16} & \dropscore{4}{20.93} & \dropscore{4}{10.20} & \dropscore{4}{4.79} & \dropscore{2}{39.19} & \dropscore{2}{2.04} \\
Pixel Unshuffle & -- & -- & -- & -- & -- & -- & -- & -- & -- & -- \\
\midrule
\rowcolor{gray!10}
\multicolumn{11}{@{}l}{\emph{Atomic methods}} \\
CLS-Attn & \dropscore{2}{51.89} & \dropscore{2}{58.11} & \dropscore{1}{51.56} & \dropscore{1}{46.00} & \dropscore{4}{15.88} & \dropscore{4}{23.91} & \dropscore{4}{11.42} & \dropscore{4}{4.94} & \dropscore{2}{38.30} & \dropscore{2}{2.00} \\
CLS-Sim & \dropscore{2}{49.15} & \dropscore{3}{52.44} & \dropscore{2}{49.56} & \dropscore{1}{45.44} & \dropscore{4}{13.93} & \dropscore{4}{20.87} & \dropscore{4}{9.78} & \dropscore{4}{4.58} & \dropscore{3}{33.79} & \dropscore{3}{1.77} \\
Bipartite Prune & \dropscore{1}{52.33} & \dropscore{2}{60.11} & \dropscore{2}{50.67} & 46.22 & \dropscore{4}{14.16} & \dropscore{4}{21.11} & \dropscore{4}{10.20} & \dropscore{4}{4.70} & \dropscore{1}{42.13} & \dropscore{1}{2.19} \\
Bipartite Merge & \dropscore{1}{53.30} & \dropscore{2}{61.44} & \dropscore{1}{52.00} & 46.44 & \dropscore{4}{14.75} & \dropscore{4}{22.18} & \dropscore{4}{10.44} & \dropscore{4}{4.76} & \dropscore{1}{41.58} & \dropscore{1}{2.15} \\
Temporal Prune & \dropscore{1}{53.37} & \dropscore{2}{60.00} & \dropscore{1}{52.33} & 47.78 & \dropscore{4}{14.86} & \dropscore{4}{22.24} & \dropscore{4}{10.56} & \dropscore{4}{4.94} & \dropscore{1}{42.10} & \dropscore{1}{2.19} \\
Selective Slowfast & \dropscore{2}{51.67} & \dropscore{2}{58.44} & \dropscore{2}{49.11} & 47.44 & \dropscore{4}{14.05} & \dropscore{4}{20.96} & \dropscore{4}{10.05} & \dropscore{4}{4.67} & \dropscore{1}{40.82} & \dropscore{1}{2.12} \\
GDR & \dropscore{1}{53.41} & \dropscore{2}{61.89} & \dropscore{1}{51.89} & 46.44 & \dropscore{4}{16.79} & \dropscore{4}{24.77} & \dropscore{4}{12.01} & \dropscore{4}{5.47} & 42.36 & \dropscore{1}{2.20} \\
FPS & \dropscore{1}{52.85} & \dropscore{2}{60.56} & \dropscore{1}{51.67} & 46.33 & \dropscore{4}{14.70} & \dropscore{4}{21.97} & \dropscore{4}{10.53} & \dropscore{4}{4.91} & 42.28 & \dropscore{1}{2.19} \\
Text-Sim & \dropscore{2}{49.37} & \dropscore{3}{53.67} & \dropscore{2}{48.33} & 46.11 & \dropscore{4}{13.98} & \dropscore{4}{20.79} & \dropscore{4}{10.11} & \dropscore{4}{4.85} & \dropscore{3}{34.32} & \dropscore{3}{1.79} \\
\midrule
\rowcolor{gray!10}
\multicolumn{11}{@{}l}{\emph{Mixed methods}} \\
CLS-Attn+Bip. Prune & \dropscore{2}{51.63} & \dropscore{2}{56.78} & \dropscore{1}{51.22} & 46.89 & \dropscore{4}{15.68} & \dropscore{4}{23.19} & \dropscore{4}{11.24} & \dropscore{4}{5.11} & \dropscore{1}{40.43} & \dropscore{1}{2.11} \\
Text-Sim+FPS & \dropscore{1}{52.22} & \dropscore{2}{58.67} & \dropscore{1}{50.89} & 47.11 & \dropscore{4}{14.45} & \dropscore{4}{21.71} & \dropscore{4}{10.26} & \dropscore{4}{4.79} & \dropscore{1}{41.11} & \dropscore{1}{2.13} \\
\bottomrule
\end{tabularx}
\end{adjustbox}
\caption{Training-free results on Nano-9B-v2.}
\label{tab:results_nano_tf}
\end{table*}

\begin{table*}[tp]
\centering
\small
\setlength{\tabcolsep}{1.6pt}
\begin{adjustbox}{max width=\textwidth}
\begin{tabularx}{\textwidth}{@{}l@{\hspace{8pt}}|*{4}{>{\centering\arraybackslash}X}|*{4}{>{\centering\arraybackslash}X}|*{2}{>{\centering\arraybackslash}X}@{}}
\toprule
\multirow{2}{*}{Method} & \multicolumn{4}{c|}{VideoMME} & \multicolumn{4}{c|}{Charades} & \multicolumn{2}{c}{VDC} \\
& Avg. & Short & Medium & Long & mIoU & R@.3 & R@.5 & R@.7 & Acc. & Score \\
\midrule
Uncompressed & 57.89 & 68.78 & 56.44 & 48.44 & 32.50 & 48.74 & 23.55 & 10.14 & 44.48 & 2.32 \\
\midrule
\rowcolor{gray!10}
\multicolumn{11}{@{}l}{\emph{Simple methods}} \\
Uniform Prune & \dropscore{1}{54.41} & \dropscore{2}{61.67} & 53.78 & 47.78 & \dropscore{2}{27.10} & \dropscore{2}{40.50} & \dropscore{2}{19.36} & \dropscore{2}{8.30} & 43.42 & 2.25 \\
Window Prune & \dropscore{1}{54.63} & \dropscore{1}{62.78} & \dropscore{1}{53.11} & 48.00 & \dropscore{2}{28.72} & \dropscore{2}{42.73} & \dropscore{2}{20.73} & \dropscore{2}{8.86} & 42.46 & 2.21 \\
Spatial Pool & 55.59 & \dropscore{1}{63.11} & 55.56 & 48.11 & \dropscore{1}{29.74} & \dropscore{1}{44.63} & \dropscore{2}{21.08} & \dropscore{1}{9.34} & 42.95 & 2.22 \\
Spatiotemporal Pool & 56.04 & \dropscore{1}{64.11} & 54.44 & 49.56 & \dropscore{1}{29.76} & \dropscore{1}{44.37} & \dropscore{2}{21.14} & \dropscore{1}{9.57} & 43.24 & 2.23 \\
Pixel Unshuffle & 56.63 & 65.78 & 56.89 & 47.22 & \dropscore{1}{30.86} & \dropscore{1}{46.15} & \dropscore{1}{22.06} & \dropscore{1}{9.52} & 43.41 & 2.26 \\
\midrule
\rowcolor{gray!10}
\multicolumn{11}{@{}l}{\emph{Atomic methods}} \\
CLS-Attn & 55.78 & \dropscore{1}{63.11} & 55.00 & 49.22 & \dropscore{1}{29.67} & \dropscore{1}{44.22} & \dropscore{1}{21.74} & \dropscore{1}{9.63} & \dropscore{1}{41.70} & \dropscore{1}{2.18} \\
CLS-Sim & \dropscore{1}{53.19} & \dropscore{2}{58.11} & \dropscore{1}{53.11} & 48.33 & \dropscore{2}{26.19} & \dropscore{2}{39.28} & \dropscore{3}{18.67} & \dropscore{2}{8.24} & \dropscore{2}{39.84} & \dropscore{2}{2.08} \\
Bipartite Prune & \dropscore{1}{53.85} & \dropscore{2}{61.33} & \dropscore{1}{53.44} & 46.78 & \dropscore{2}{28.17} & \dropscore{2}{41.90} & \dropscore{2}{19.77} & \dropscore{2}{8.36} & 43.46 & 2.25 \\
Bipartite Merge & \dropscore{1}{54.44} & \dropscore{1}{63.44} & \dropscore{1}{51.56} & 48.33 & \dropscore{1}{30.28} & \dropscore{1}{45.52} & \dropscore{1}{21.94} & \dropscore{1}{9.63} & 44.45 & 2.29 \\
Temporal Prune & \dropscore{1}{53.56} & \dropscore{2}{60.56} & \dropscore{1}{52.56} & 47.56 & \dropscore{3}{25.75} & \dropscore{2}{39.28} & \dropscore{3}{18.70} & \dropscore{3}{7.85} & 43.10 & 2.24 \\
Selective Slowfast & \dropscore{1}{53.81} & \dropscore{2}{60.67} & \dropscore{1}{52.78} & 48.00 & \dropscore{3}{25.97} & \dropscore{3}{38.69} & \dropscore{3}{18.47} & \dropscore{3}{7.97} & 43.04 & 2.23 \\
GDR & \dropscore{1}{54.44} & \dropscore{1}{62.11} & 54.44 & 46.78 & \dropscore{1}{30.48} & \dropscore{1}{46.00} & \dropscore{1}{21.80} & \dropscore{1}{9.31} & \dropscore{1}{41.61} & \dropscore{1}{2.17} \\
FPS & \dropscore{1}{54.70} & \dropscore{1}{65.11} & \dropscore{1}{52.56} & 46.44 & \dropscore{2}{28.72} & \dropscore{2}{43.21} & \dropscore{2}{20.87} & \dropscore{1}{9.55} & 43.18 & 2.25 \\
Text-Sim & \dropscore{2}{51.63} & \dropscore{2}{55.67} & \dropscore{1}{52.00} & 47.22 & \dropscore{2}{26.24} & \dropscore{2}{39.52} & \dropscore{3}{18.76} & \dropscore{3}{8.06} & \dropscore{1}{41.10} & \dropscore{1}{2.15} \\
\midrule
\rowcolor{gray!10}
\multicolumn{11}{@{}l}{\emph{Mixed methods}} \\
CLS-Attn+Bip. Prune & 55.07 & \dropscore{1}{63.56} & 54.00 & 47.67 & \dropscore{2}{29.06} & \dropscore{2}{43.68} & \dropscore{2}{20.73} & \dropscore{2}{8.83} & 43.14 & 2.23 \\
Text-Sim+FPS & \dropscore{1}{54.11} & \dropscore{1}{64.22} & \dropscore{1}{52.89} & \dropscore{1}{45.22} & \dropscore{2}{28.96} & \dropscore{2}{43.47} & \dropscore{2}{20.99} & \dropscore{1}{9.34} & 42.60 & 2.21 \\
\bottomrule
\end{tabularx}
\end{adjustbox}
\caption{Fine-tuned results on Nano-9B-v2.}
\label{tab:results_nano_ft}
\end{table*}

\begin{table*}[tp]
\centering
\small
\setlength{\tabcolsep}{4pt}
\begin{adjustbox}{max width=\textwidth}
\begin{tabular}{@{}l | rrl | rr | rr | rrl@{}}
\toprule
\multirow{2}{*}{Method} & \multicolumn{3}{c|}{\shortstack{Prefill Time\\(ms) $\downarrow$}} & \multicolumn{2}{c|}{\shortstack{Throughput\\(tokens/s) $\uparrow$}} & \multicolumn{2}{c|}{\shortstack{Compress. Time\\(ms) $\downarrow$}} & \multicolumn{3}{c}{\shortstack{Peak Memory\\(GiB) $\downarrow$}} \\
 & 32frm & 256frm & \textit{Scale} & 32frm & 256frm & 32frm & 256frm & 32frm & 256frm & \textit{Scale} \\
\midrule
Uncompressed & 276.8 & 4201.6 & \textcolor{gray}{(15.2$\times$)} & 52.7 & 46.8 & 0.0 & 0.0 & 18.1 & 39.3 & \textcolor{gray}{(2.2$\times$)} \\
\midrule
\multicolumn{11}{l}{\textit{Projector-level compression ($16\times$)}} \\
Uniform Prune & 54.5 & 306.6 & \textcolor{gray}{(5.6$\times$)} & 52.1 & 52.2 & 0.6 & 4.0 & 15.4 & 17.4 & \textcolor{gray}{(1.1$\times$)} \\
Window Prune & 55.4 & 306.9 & \textcolor{gray}{(5.5$\times$)} & 50.4 & 51.7 & 0.7 & 3.9 & 15.4 & 17.4 & \textcolor{gray}{(1.1$\times$)} \\
Spatial Pool & 55.5 & 303.4 & \textcolor{gray}{(5.5$\times$)} & 49.5 & 49.3 & 0.3 & 0.5 & 15.4 & 17.4 & \textcolor{gray}{(1.1$\times$)} \\
Spatiotemporal Pool & 56.4 & 313.2 & \textcolor{gray}{(5.6$\times$)} & 50.9 & 50.3 & 1.5 & 10.3 & 15.4 & 17.4 & \textcolor{gray}{(1.1$\times$)} \\
CLS-Attn & 55.9 & 304.4 & \textcolor{gray}{(5.4$\times$)} & 52.1 & 52.3 & 0.6 & 1.2 & 15.4 & 17.8 & \textcolor{gray}{(1.2$\times$)} \\
CLS-Sim & 54.0 & 302.4 & \textcolor{gray}{(5.6$\times$)} & 52.0 & 50.6 & 0.4 & 1.0 & 15.3 & 16.8 & \textcolor{gray}{(1.1$\times$)} \\
Bipartite Prune & 55.2 & 307.3 & \textcolor{gray}{(5.6$\times$)} & 52.7 & 52.9 & 1.9 & 3.7 & 15.4 & 17.8 & \textcolor{gray}{(1.2$\times$)} \\
Bipartite Merge & 58.4 & 317.8 & \textcolor{gray}{(5.4$\times$)} & 50.9 & 49.5 & 4.3 & 13.4 & 15.4 & 17.8 & \textcolor{gray}{(1.2$\times$)} \\
Temporal Prune & 56.5 & 321.2 & \textcolor{gray}{(5.7$\times$)} & 51.5 & 50.8 & 2.6 & 18.3 & 15.4 & 17.6 & \textcolor{gray}{(1.1$\times$)} \\
Selective SlowFast & 56.2 & 313.5 & \textcolor{gray}{(5.6$\times$)} & 52.1 & 52.6 & 1.5 & 9.3 & 15.4 & 17.5 & \textcolor{gray}{(1.1$\times$)} \\
GDR & 136.1 & 4645.7 & \textcolor{gray}{(34.1$\times$)} & 52.4 & 53.9 & 83.4 & 4452.9 & 15.6 & 19.4 & \textcolor{gray}{(1.2$\times$)} \\
FPS & 225.6 & 8878.0 & \textcolor{gray}{(39.4$\times$)} & 53.0 & 52.2 & 174.2 & 8578.6 & 15.6 & 19.0 & \textcolor{gray}{(1.2$\times$)} \\
Text-Attn & 57.1 & 348.2 & \textcolor{gray}{(6.1$\times$)} & 52.3 & 52.6 & 5.0 & 47.5 & 15.7 & 34.1 & \textcolor{gray}{(2.2$\times$)} \\
Text-Sim & 54.6 & 306.1 & \textcolor{gray}{(5.6$\times$)} & 53.5 & 52.9 & 3.0 & 5.1 & 15.5 & 18.2 & \textcolor{gray}{(1.2$\times$)} \\
CLS-Attn+Bip. Prune & 54.9 & 307.4 & \textcolor{gray}{(5.6$\times$)} & 53.0 & 53.4 & 1.9 & 3.4 & 15.4 & 17.8 & \textcolor{gray}{(1.2$\times$)} \\
Text-Sim+FPS & 144.2 & 4598.2 & \textcolor{gray}{(31.9$\times$)} & 50.0 & 51.5 & 89.3 & 4301.4 & 15.6 & 19.0 & \textcolor{gray}{(1.2$\times$)} \\
Text-Attn+FPS & 141.9 & 4629.5 & \textcolor{gray}{(32.6$\times$)} & 53.4 & 52.9 & 91.4 & 4336.3 & 15.8 & 34.1 & \textcolor{gray}{(2.2$\times$)} \\
\midrule
\multicolumn{11}{l}{\textit{LLM-internal compression (L1)}} \\
Uniform Prune & 58.9 & 369.9 & \textcolor{gray}{(6.3$\times$)} & 51.7 & 51.7 & 2.0 & 3.7 & 16.3 & 24.7 & \textcolor{gray}{(1.5$\times$)} \\
CLS-Attn & 58.2 & 373.2 & \textcolor{gray}{(6.4$\times$)} & 54.2 & 53.9 & 2.3 & 5.3 & 16.4 & 25.7 & \textcolor{gray}{(1.6$\times$)} \\
CLS-Sim & 58.3 & 370.1 & \textcolor{gray}{(6.3$\times$)} & 51.2 & 50.7 & 2.3 & 4.0 & 16.3 & 24.7 & \textcolor{gray}{(1.5$\times$)} \\
Bipartite Prune & 59.8 & 380.1 & \textcolor{gray}{(6.4$\times$)} & 52.9 & 52.6 & 3.9 & 12.2 & 16.4 & 25.7 & \textcolor{gray}{(1.6$\times$)} \\
Bipartite Merge & 61.1 & 387.1 & \textcolor{gray}{(6.3$\times$)} & 52.9 & 53.1 & 5.5 & 20.4 & 16.4 & 25.7 & \textcolor{gray}{(1.6$\times$)} \\
Temporal Prune$\ddagger$ & 59.7 & 382.5 & \textcolor{gray}{(6.4$\times$)} & 48.9 & 51.5 & 1.8 & 15.2 & 16.3 & 24.7 & \textcolor{gray}{(1.5$\times$)} \\
Selective SlowFast & 59.5 & 377.8 & \textcolor{gray}{(6.3$\times$)} & 53.3 & 50.8 & 3.3 & 12.5 & 16.3 & 24.7 & \textcolor{gray}{(1.5$\times$)} \\
GDR & 94.5 & 2338.8 & \textcolor{gray}{(24.8$\times$)} & 53.1 & 53.0 & 41.0 & 1981.8 & 16.3 & 24.7 & \textcolor{gray}{(1.5$\times$)} \\
FPS & 134.4 & 4194.7 & \textcolor{gray}{(31.2$\times$)} & 53.0 & 50.0 & 78.8 & 3830.4 & 16.3 & 24.7 & \textcolor{gray}{(1.5$\times$)} \\
Text-Attn & 61.1 & 412.2 & \textcolor{gray}{(6.7$\times$)} & 52.7 & 51.5 & 5.4 & 48.7 & 16.3 & 34.7 & \textcolor{gray}{(2.1$\times$)} \\
Text-Sim & 60.6 & 371.2 & \textcolor{gray}{(6.1$\times$)} & 50.4 & 51.1 & 2.8 & 4.4 & 16.3 & 24.7 & \textcolor{gray}{(1.5$\times$)} \\
CLS-Attn+Bip. Prune & 59.9 & 380.7 & \textcolor{gray}{(6.4$\times$)} & 53.8 & 53.7 & 3.7 & 11.9 & 16.4 & 25.7 & \textcolor{gray}{(1.6$\times$)} \\
Text-Sim+FPS & 97.0 & 2289.4 & \textcolor{gray}{(23.6$\times$)} & 52.8 & 49.4 & 41.7 & 1927.2 & 16.3 & 24.7 & \textcolor{gray}{(1.5$\times$)} \\
Text-Attn+FPS & 99.7 & 2321.9 & \textcolor{gray}{(23.3$\times$)} & 53.1 & 53.0 & 43.3 & 1958.7 & 16.3 & 34.7 & \textcolor{gray}{(2.1$\times$)} \\
Text-Attn$\dagger$ & 60.2 & 404.4 & \textcolor{gray}{(6.7$\times$)} & 52.8 & 53.3 & 4.1 & 39.6 & 16.3 & 34.7 & \textcolor{gray}{(2.1$\times$)} \\
\bottomrule
\end{tabular}
\end{adjustbox}
\caption{Qwen efficiency results.}
\label{tab:efficiency_qwen}
\end{table*}

\begin{table*}[tp]
\centering
\small
\setlength{\tabcolsep}{4pt}
\begin{adjustbox}{max width=\linewidth}
\begin{tabular}{@{}l | rrl | rr | rr | rrl@{}}
\toprule
\multirow{2}{*}{Method} & \multicolumn{3}{c|}{\shortstack{Prefill Time\\(ms) $\downarrow$}} & \multicolumn{2}{c|}{\shortstack{Decoding\\(tokens/s) $\uparrow$}} & \multicolumn{2}{c|}{\shortstack{Compress. Time\\(ms) $\downarrow$}} & \multicolumn{3}{c}{\shortstack{Peak Memory\\(GiB) $\downarrow$}} \\
 & 32frm & 256frm & \textit{Scale} & 32frm & 256frm & 32frm & 256frm & 32frm & 256frm & \textit{Scale} \\
\midrule
Uncompressed & 303.6 & 2799.1 & \textcolor{gray}{(9.2$\times$)} & 32.6 & 26.8 & 0.0 & 0.0 & 24.1 & 69.2 & \textcolor{gray}{(2.9$\times$)} \\
\midrule
\multicolumn{11}{l}{\textit{Projector-level compression ($16\times$)}} \\
Uniform Prune & 81.9 & 341.7 & \textcolor{gray}{(4.2$\times$)} & 34.0 & 33.9 & 3.8 & 14.1 & 18.3 & 22.7 & \textcolor{gray}{(1.2$\times$)} \\
Window Prune & 82.4 & 335.4 & \textcolor{gray}{(4.1$\times$)} & 34.2 & 34.2 & 0.8 & 4.2 & 18.1 & 21.7 & \textcolor{gray}{(1.2$\times$)} \\
Spatial Pool & 84.4 & 332.2 & \textcolor{gray}{(3.9$\times$)} & 32.0 & 33.3 & 0.2 & 0.7 & 18.1 & 21.7 & \textcolor{gray}{(1.2$\times$)} \\
Spatiotemporal Pool & 84.7 & 340.8 & \textcolor{gray}{(4.0$\times$)} & 32.2 & 34.3 & 1.5 & 10.9 & 18.1 & 21.7 & \textcolor{gray}{(1.2$\times$)} \\
CLS-Attn & 80.3 & 330.4 & \textcolor{gray}{(4.1$\times$)} & 34.0 & 34.3 & 0.5 & 1.4 & 18.2 & 22.0 & \textcolor{gray}{(1.2$\times$)} \\
CLS-Sim & 83.0 & 329.4 & \textcolor{gray}{(4.0$\times$)} & 32.6 & 34.1 & 0.4 & 0.9 & 18.0 & 21.0 & \textcolor{gray}{(1.2$\times$)} \\
Bipartite Prune & 80.4 & 333.1 & \textcolor{gray}{(4.1$\times$)} & 34.7 & 34.5 & 1.8 & 3.4 & 18.2 & 22.0 & \textcolor{gray}{(1.2$\times$)} \\
Bipartite Merge & 81.3 & 341.5 & \textcolor{gray}{(4.2$\times$)} & 34.9 & 34.9 & 4.3 & 13.3 & 18.2 & 22.0 & \textcolor{gray}{(1.2$\times$)} \\
Temporal Prune & 82.0 & 349.7 & \textcolor{gray}{(4.3$\times$)} & 34.8 & 33.4 & 2.8 & 19.7 & 18.1 & 21.7 & \textcolor{gray}{(1.2$\times$)} \\
Selective SlowFast & 81.2 & 341.7 & \textcolor{gray}{(4.2$\times$)} & 34.6 & 34.6 & 1.7 & 10.0 & 18.1 & 21.7 & \textcolor{gray}{(1.2$\times$)} \\
GDR & 166.7 & 4997.2 & \textcolor{gray}{(30.0$\times$)} & 33.7 & 34.0 & 88.4 & 4653.3 & 18.3 & 22.8 & \textcolor{gray}{(1.2$\times$)} \\
FPS & 285.4 & 10923.1 & \textcolor{gray}{(38.3$\times$)} & 34.8 & 34.5 & 207.5 & 10593.5 & 18.3 & 22.7 & \textcolor{gray}{(1.2$\times$)} \\
Text-Sim & 82.1 & 333.4 & \textcolor{gray}{(4.1$\times$)} & 33.7 & 34.8 & 2.8 & 5.3 & 18.3 & 22.7 & \textcolor{gray}{(1.2$\times$)} \\
CLS-Attn+Bip. Prune & 81.6 & 333.4 & \textcolor{gray}{(4.1$\times$)} & 34.5 & 34.5 & 2.2 & 4.4 & 18.2 & 22.0 & \textcolor{gray}{(1.2$\times$)} \\
Text-Sim+FPS & 185.7 & 5640.4 & \textcolor{gray}{(30.4$\times$)} & 33.8 & 34.1 & 106.8 & 5311.9 & 18.3 & 22.7 & \textcolor{gray}{(1.2$\times$)} \\
\midrule
\multicolumn{11}{l}{\textit{LLM-internal compression at L2 (target $36\times$)}} \\
Uniform Prune & 82.9 & 345.4 & \textcolor{gray}{(4.2$\times$)} & 34.1 & 33.6 & 3.5 & 12.5 & 18.9 & 28.2 & \textcolor{gray}{(1.5$\times$)} \\
CLS-Attn & 85.9 & 345.2 & \textcolor{gray}{(4.0$\times$)} & 32.0 & 30.2 & 2.3 & 5.8 & 19.0 & 29.2 & \textcolor{gray}{(1.5$\times$)} \\
CLS-Sim & 81.5 & 336.8 & \textcolor{gray}{(4.1$\times$)} & 34.6 & 34.3 & 2.4 & 4.4 & 18.9 & 28.2 & \textcolor{gray}{(1.5$\times$)} \\
Bipartite Prune & 84.3 & 348.0 & \textcolor{gray}{(4.1$\times$)} & 33.7 & 33.9 & 3.8 & 13.9 & 19.0 & 29.2 & \textcolor{gray}{(1.5$\times$)} \\
Bipartite Merge & 85.7 & 357.8 & \textcolor{gray}{(4.2$\times$)} & 34.3 & 34.6 & 6.3 & 23.9 & 19.0 & 29.2 & \textcolor{gray}{(1.5$\times$)} \\
Temporal Prune$\ddagger$ & 83.5 & 348.1 & \textcolor{gray}{(4.2$\times$)} & 32.7 & 34.3 & 1.7 & 16.6 & 18.9 & 28.2 & \textcolor{gray}{(1.5$\times$)} \\
Selective SlowFast & 85.3 & 345.4 & \textcolor{gray}{(4.0$\times$)} & 33.9 & 34.4 & 3.5 & 13.7 & 18.9 & 28.2 & \textcolor{gray}{(1.5$\times$)} \\
GDR & 121.1 & 2457.5 & \textcolor{gray}{(20.3$\times$)} & 34.1 & 34.1 & 42.1 & 2127.8 & 18.9 & 28.2 & \textcolor{gray}{(1.5$\times$)} \\
FPS & 173.7 & 5046.1 & \textcolor{gray}{(29.0$\times$)} & 34.4 & 34.1 & 94.1 & 4716.1 & 18.9 & 28.2 & \textcolor{gray}{(1.5$\times$)} \\
Text-Sim & 82.1 & 336.8 & \textcolor{gray}{(4.1$\times$)} & 34.3 & 34.3 & 2.6 & 4.7 & 18.9 & 28.2 & \textcolor{gray}{(1.5$\times$)} \\
CLS-Attn+Bip. Prune & 84.3 & 347.8 & \textcolor{gray}{(4.1$\times$)} & 34.2 & 34.0 & 3.8 & 13.5 & 19.0 & 29.2 & \textcolor{gray}{(1.5$\times$)} \\
Text-Sim+FPS & 129.1 & 2702.3 & \textcolor{gray}{(20.9$\times$)} & 33.9 & 32.7 & 48.8 & 2367.8 & 18.9 & 28.2 & \textcolor{gray}{(1.5$\times$)} \\
\bottomrule
\end{tabular}
\end{adjustbox}
\caption{Nano efficiency results.}
\label{tab:efficiency_nano}
\end{table*}

\end{document}